%% file: neurips_2026.tex
\documentclass{article}

\PassOptionsToPackage{numbers, compress}{natbib}

\usepackage[dblblindworkshop, final]{neurips_2026}
\makeatletter
\renewcommand{\@noticestring}{TAE (Trust-AI-Eval): Can We Trust AI Evaluation?}
\makeatother

\workshoptitle{TAE (Trust-AI-Eval): Can We Trust AI Evaluation?}

\usepackage[utf8]{inputenc} 
\usepackage[T1]{fontenc}    
\usepackage{hyperref}       
\usepackage{url}            
\usepackage{booktabs}       
\usepackage{amsfonts}       
\usepackage{nicefrac}       
\usepackage{microtype}      
\usepackage{xcolor}         

\usepackage{xspace}

\usepackage{amsmath}
\usepackage{cleveref}
\usepackage{enumitem}
\usepackage{booktabs}
\usepackage{multirow}
\usepackage[textsize=tiny]{todonotes}

\usepackage{mdframed}
\usepackage{amsthm}
\theoremstyle{definition} 
\usepackage{subcaption}

\usepackage{etoolbox}

\newmdtheoremenv[%
linecolor=gray,leftmargin=10,%
rightmargin=10,
backgroundcolor=gray!40,
nobreak=false,
]{myprop}{Template}[section]

\usepackage{listings}
\definecolor{mycyan}{RGB}{155, 253, 234}
\definecolor{myyellow}{RGB}{255,252,102}
\definecolor{mygreen}{RGB}{136, 250, 78}
\definecolor{mypurple}{RGB}{190, 190, 254}
\definecolor{mypink}{RGB}{255,141,198}
\definecolor{myblue}{RGB}{86,193,255}

\newcommand{\methodname}{\textsc{Comprehend}\xspace}
\newcommand{\heading}[1]{\vspace{5pt}\noindent\underline{\textsc{#1}}}

\title{\methodname: Ultra-Fine-Grained Benchmark Understanding by Entailment Hierarchy}

\author{%
Sivaraaman Balakrishnan\thanks{\, indicates equal contribution.}\;\textsuperscript{1} \;\; Prakriti Shetty\footnotemark[1]\;\textsuperscript{1} \;\; Jeremiah Hong\textsuperscript{1} \\ 
\textbf{Hamed Zamani\textsuperscript{1}} \;\; \textbf{Haw-Shiuan Chang\footnotemark[1]\;\textsuperscript{2}} \\
        \textsuperscript{1}University of Massachusetts, Amherst, USA \\
        \textsuperscript{2}University of Arizona, Tucson, USA \\
\texttt{ \{psshetty,sivaraamanba,jshong\}@umass.edu}, \\ \texttt{ zamani@cs.umass.edu, hawshiuan@arizona.edu}
}

\begin{document}

\maketitle

\begin{abstract}
Understanding what kind of prompts users typically ask and how well an LLM performs on these prompt categories is crucial for LLM practitioners. The existing methods satisfy the demand by hierarchically clustering the user prompts and synthesizing the cluster names using an LLM to build a taxonomy. However, the cluster names from these methods are usually too general for humans to understand the detailed prompt distributions and identify the exact strengths and weaknesses of LLMs. In this study, we propose \methodname, which automatically constructs an ultra-fine-grained entailment hierarchy for the prompts. The entailment hierarchy replaces the cluster names in the taxonomy with sentences to provide more precise cluster descriptions for more fine-grained categorization. Compared with EvalTree, a state-of-the-art taxonomy construction method, \methodname improves an entailment/faithfulness metric from $22.6$ to $41.7$ and an informative metric from $16.3$ to $39.1$ in WildBench. Our experiments also show that the weakness report generated based on our entailment hierarchy is much more informative than EvalTree in terms of predicting the response scores in the benchmark. All the code is released in
\url{https://github.com/JackNapier20/COMPREHEND_public}.
\end{abstract}

\input{contents/intro}

\input{contents/related_work}

\input{contents/methods}

\input{contents/experiments}

\input{contents/conclusion_limitation}

\section*{Acknowledgments}
This work was supported in part by the Center for Intelligent Information Retrieval, in part by the National Science Foundation grant \#2106391, and in part by Cisco. Any opinions, findings and conclusions or recommendations expressed in this material are those of the authors and do not necessarily reflect those of the sponsor.

\bibliographystyle{abbrvnat}
\bibliography{custom}


\appendix

\clearpage

\input{appendix/more_results}

\input{appendix/limitations}

\input{appendix/method_details}

\input{appendix/experiment_details}



\end{document}

%% file: contents/intro.tex
\section{Introduction}


A great tool that can analyze a large number of prompts and LLMs' performances on these prompts is demanded by many LLM practitioners. For instance, product managers or safety researchers sometimes have to monitor millions of user prompts to identify users' needs or discover misuse patterns~\cite{tamkin2024clio}. One daily job of LLM developers is using benchmarks to discover the weaknesses or blind spots of LLMs, compare their performances with the competitors, and understand the effect of training strategies. Moreover, many LLM users want to select the LLMs for their tasks using more fine-grained LLM characteristics than the categories such as math reasoning and creative writing from existing benchmarks. 

To address this challenge, a common approach is hierarchically clustering (the summary of) the prompts and asking an LLM to generate the category name of each cluster~\cite{tamkin2024clio,cheng2024autodetect, zeng2025evaltree,wang2026scan}. The resulting taxonomy has many applications. For example, the approach could cluster the prompts about \textit{neural network} together and if the average performance of the target LLM in this cluster is worse than competing LLMs, the developer could identify the weakness and mine the related training data to fix the weakness in the next round.

\begin{figure}[t]
  \includegraphics[width=\linewidth]{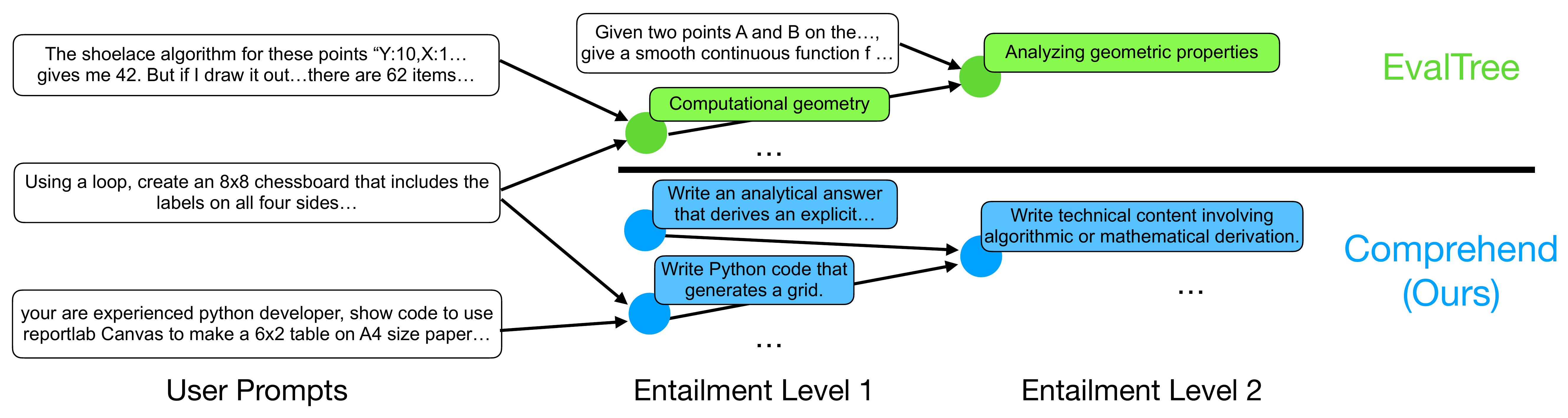}
  \caption{Comparison between EvalTree~\citep{zeng2025evaltree} and \methodname. Given the prompt of \textit{Creating a chess board using python} in the middle, \methodname groups it with another more relevant prompt and generates a more specific prompt as the group name.}
  \label{fig:compare_first}
\end{figure}

Although widely used, the approach has four major limitations. First, the synthesized category names are usually coarse-grained phrases or very short sentences. This decreases the granularity of the weakness analyses.
Second, the resulting trees from previous approaches are often shallow or imbalanced. The number of descendants (i.e., cluster size) of a tree node at a higher level could be much smaller than the node at a lower level and have a much more specific cluster name. 
Third, clustering and category name generation are often two separate steps in their pipeline, so clustering could group very diverse prompts together, which makes it hard for LLMs to generate any informative category name. Finally, previous studies lack intrinsic metrics to compare different tree construction methods without the ground truth of clustering.



To address the limitations, we propose \methodname (\textbf{COM}mon \textbf{PR}ompt \textbf{E}ntailment \textbf{H}ierarchy for \textbf{E}valuation a\textbf{N}d \textbf{D}evelopment). \Cref{fig:compare_first} highlights the differences between \methodname and EvalTree~\citep{zeng2025evaltree}, a representative state-of-the-art approach. 
Unlike a typical taxonomy from EvalTree, where each category is represented by a short phrase and an arrow means the child node belongs to the parent node, the nodes in the entailment hierarchy are represented by a sentence or even a passage, and an arrow means the child node entails the parent node. 

The entailment hierarchy is constructed bottom up, and the nodes in the lowest level are the user prompts in the benchmark of interest. At each level, we first use sentence BERT~\citep{reimers2019sentence} and the Blossom algorithm~\citep{gabow1990data} to match similar prompts in pairs. Then, we ask an LLM to generate the common summary of two similar prompts as a node in the next level and repeat the process recursively. 
By interleaving the generation and clustering at each level, we reduce the difficulty of generating the specific category description for the LLM, which results in a more fine-grained hierarchy.

Comparing different prompt categorization methods is challenging because there is usually no ground truth clustering annotations and different methods could have very different hierarchical structures. We propose several low-cost intrinsic metrics by adapting the metrics from summarization tasks and specificity scores from \citet{chang2024fine} and use them to evaluate how well the entailment hierarchy can help an LLM (as a human proxy) write a performance report. In WildBench~\citep{lin2025wildbench} and UltraFeedback~\citep{ultrafeedback}, we demonstrate each component of \methodname makes the cluster descriptions significantly more faithful and informative summaries of the prompts in the cluster.


\heading{Main Contributions}
\begin{itemize}[itemsep=2pt, parsep=0pt, topsep=1pt,leftmargin=*]
    \item We derive ultra-fine-grained categories of prompts using an entailment hierarchy and propose a novel bottom-up construction method \methodname.
    \item We propose general evaluation metrics that allow researchers to compare different tree structures and node representations. We also conduct comprehensive experiments and examine the performance of each proposed component.
    \item We release our code that allows researchers to build an entailment hierarchy from a corpus. 
\end{itemize}





\begin{figure*}[t]
  \includegraphics[width=\linewidth]{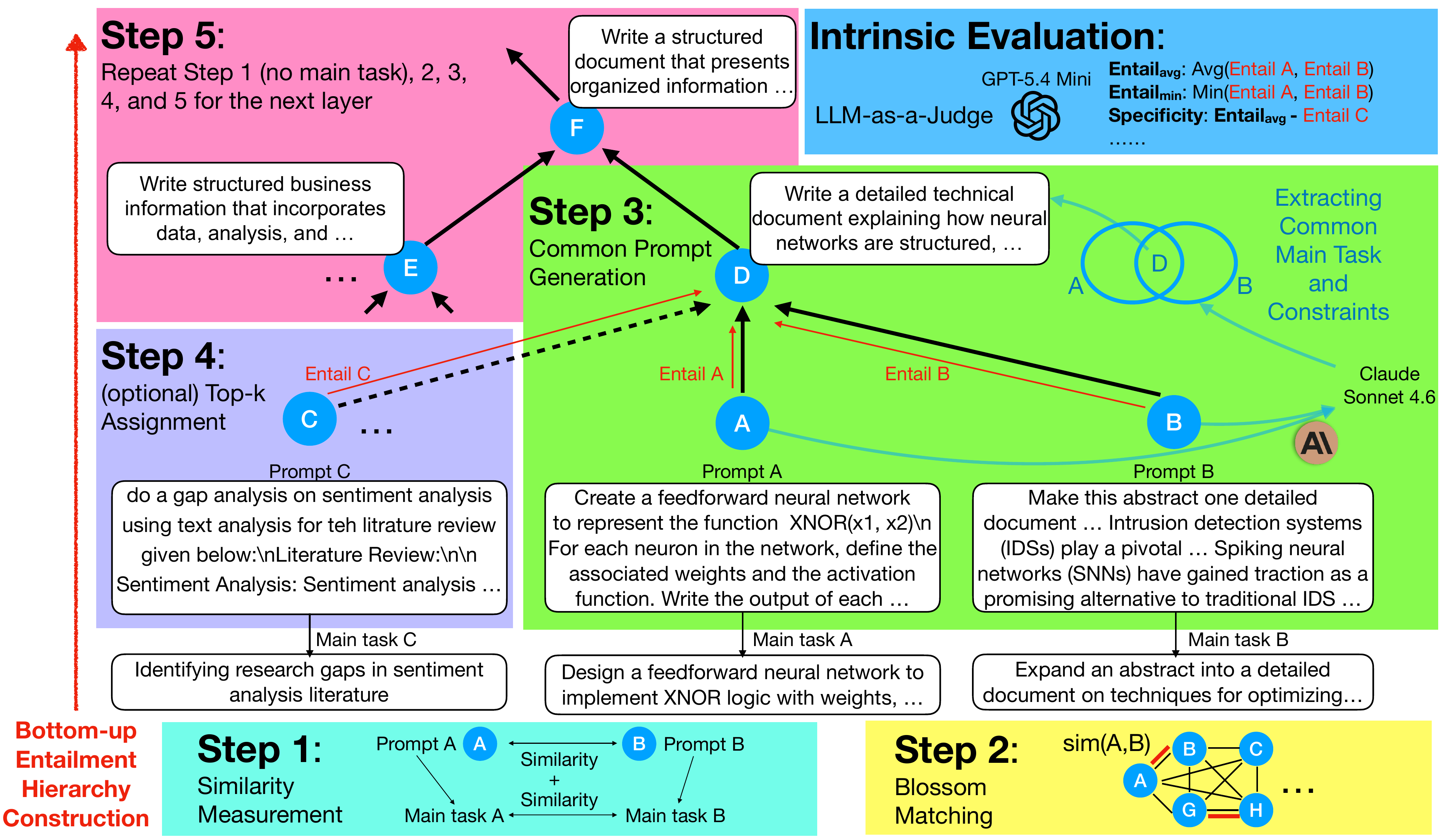}
  \caption{\fboxsep=1pt The \methodname bottom-up construction pipeline. \colorbox{mycyan}{Step 1} and \colorbox{myyellow}{Step 2}: Pairing nodes A and B using similarity measurement and Blossom matching algorithm. \colorbox{mygreen}{Step 3}: Generating the common prompt as upper-layer node D from node A and B. \colorbox{mypurple}{Step 4}: Optionally retrieve node C that is similar to A and assign it to node D if its entailment score is high. \colorbox{mypink}{Step 5}: Repeating the steps by generating the next-layer node F from the similar node E and D. \colorbox{myblue}{Intrinsic Evaluation}: We evaluate the quality of common prompt not only by the entailment score average (Entail$_{avg}$) but also by the lowest entailment score (Entail$_{min}$) to penalize only summarizing one of the prompts.} 
  \label{fig:workflow}
\end{figure*}


%% file: contents/related_work.tex
\section{Related Work}

Several prior studies build a taxonomy on prompts to facilitate our understanding of users or LLMs. The taxonomy could be constructed manually by humans~\citep{budagam2025hierarchical, jiang2026artificial} or automatically by LLMs~\cite{tamkin2024clio,cheng2024autodetect,zeng2025evaltree,wang2026scan}. The LLM-based taxonomy construction methods are also used to categorize scientific papers~\citep{hsu2024chime, kargupta2025taxoadapt, gao2025science}, social media claims~\citep{zhang2025llmtaxo}, model errors~\citep{tian2025skillverse,ashury2026errormap, huang2026probellm,kim2026evalet}, and other types of text data~\citep{tian2024generic, zhong2025hicode}. However, the resulting taxonomies are often shallow and coarse-grained.


Building the entailment or summarization structure among texts is also explored in several applications. For example, entailment tree~\citep{dalvi2021explaining, weir2024nellie, sanders2024tv} discovers the entailment relationship between multiple pieces of evidence and the claims. Building hierarchical summaries of corpus could also improve the retrieval performance~\citep{chen2023walking, sarthi2024raptor, gupta2025llm, zhang2025sirerag, huang2025retrieval, zhao2026hierarchical}. Nevertheless, their goal is not creating categories that are easily understandable to humans, so the text of one node could often include information that is not relevant to some of its descendants. 

Our motivation for finding ultra-fine-grained categories is inspired by \citet{frick2025prompt,tjuatja2025behaviorbox}, whose goal is to predict the LLM performance difference in a very specific case, but they cannot provide high-level and navigable comparison for any benchmark as we do. Methodologically, our entailment hierarchy for prompts is inspired by \citet{chang2024fine}, which proposes to use a very shallow entailment hierarchy to improve interactive story generation. Besides the motivation differences, their paper focuses on improving the efficiency of fine-tuning data collection, while our paper builds the entailment hierarchy construction system based on LLMs.



%% file: contents/methods.tex
\section{Methods}

In our common prompt entailment hierarchy, the leaves are the benchmark's user prompts. The internal nodes are sentence- to passage-length prompts, each entailed by every prompt beneath it
\citep{cattan2023champ, chang2024fine}. 
Every internal node should be entailed by all of its descendants due to the transitivity of entailment and represents a cluster of the leaf descendant prompts. 
The entailment is a generalization of the is-a relation in a taxonomy, so similarly, the nodes at the higher level are more general.

\subsection{Construction Overview}

\methodname builds the entailment hierarchy of a benchmark from bottom to top, as summarized in \Cref{fig:workflow}. At each level, we match every node to a similar partner,
summarize each pair into one \emph{common prompt} that both partners entail, and 
repeat the process at the next-level common prompts; the recursion continues until reaching the single root node. At each layer, we can optionally assign more nodes that are similar to a child node to the parent node, which augments the full binary tree into a directed acyclic graph (DAG). 
Our goal is to maximize the entailment scores of both nodes A and B (i.e., Entail A and Entail B) while keeping the common prompt specific enough so that other nodes would have a low entailment score (e.g., Entail C). The entailment score is how much the description of their parent (e.g., node D) is entailed by the lower-level node and is judged by an evaluation LLM in our experiments. 


\subsection{Matching Similar Nodes}
\label{sec:matching}

In step~1 of \Cref{fig:workflow}, we measure the pairwise similarity between every two nodes. The prompts from real users could be noisy: the actual request is sometimes buried under contexts, including background, pasted data, and formatting instructions.
To group two prompts only when both their context and their underlying task are similar, we extract a one-sentence \emph{main task} per prompt $X$, task$_X$, with an LLM at the leaf level and add its similarity to the full-text similarity:
\begin{align}
\mathrm{sim}(A,B)=\; \mathrm{sim}(\text{prompt}_A,\text{prompt}_B) + \mathrm{sim}(\text{task}_A,\text{task}_B),
\end{align}
where the similarity function $\mathrm{sim}(,)$ is computed by the cosine similarity between the embeddings of the two texts from Sentence-BERT~\citep{reimers2019sentence} (\texttt{all-mpnet-base-v2}). The internal nodes usually do not contain lots of context information, so the full-text similarity $\mathrm{sim}(\text{common prompt}_A,\text{common prompt}_B)$ alone is sufficient.


In step~2 of \Cref{fig:workflow}, we pair the nodes using the pairwise similarity. To maximize the total similarity, we formulate a maximum-weight matching problem in a complete graph $G=(V,E)$ with weights $w(i,j)=\mathrm{sim}(i,j)$ and all nodes at a level $V$. 
A matching $M\subseteq E$ is a set of vertex-disjoint edges with maximum size (i.e., $|M| = \lfloor \frac{|V|}{2}  \rfloor$); we take the one of greatest total weight among all matchings,
\begin{align}
M^\star=\arg\max_{M\in\mathcal{M}}\ \sum_{(i,j)\in M}\mathrm{sim}(i,j),
\end{align}
which Blossom algorithm \citep{gabow1990data} could solve exactly in $O\left(|V|( |E| + |V| \log |V|) \right)$. 






\subsection{Common Prompt Generation}
\label{sec:common_prompt}
In step~3 of \Cref{fig:workflow}, our goal is to generate a specific common prompt of the two similar prompts matched in steps 1 and 2. Given a prompt pair $(A,B)$, an LLM writes a common prompt $D$ that is as specific as possible and generalizes both $A$ and $B$. $D$ is their intersection, not their union: it may state only what $A$ and $B$ both ask for, and nothing particular to one of them. 
In the example of \Cref{fig:workflow}, prompt $A$ \textit{implements an XNOR network} and $B$ expands an abstract on intrusion detection; their common prompt asks to ``\textit{write a detailed technical document explaining how neural networks are structured},'' not the union ``\textit{create a feed-forward neural network and extend an abstract},'' which would carry constraints that neither prompt demands of the other. 

The intersection requirement is what makes entailment transitive along the tree. Once $A$ and $B$
both entail $D$, the edges $A\!\to\!D$ and $B\!\to\!D$ hold; imposed at every node, this closes the
relation so that a leaf entails not only its parent but every ancestor up to the root. A union would
break the closure: a parent that kept a detail unique to $A$ would no longer be entailed by $B$,
nor by anything beneath $B$. 
We discard and regenerate a common prompt whenever its lowest child entailment score falls below the threshold $0.4$, for up to two retries.


\subsection{Optional Assignment}
\label{sec:optional}
Step 2 and step 3 leave two potential problems. First, if the number of nodes $|V|$ in a layer is odd, there must be one unmatched node left in step 2. Second, some prompt nodes could entail multiple nodes in the upper layer, but step 3 allows each node to have only one parent. 

To solve the problems, we create more entailment arrows at each layer, which allows the entailment hierarchy to be a directed acyclic graph (DAG) rather than a binary tree. First, when $|V|$ is odd at each layer, we assign the unmatched node to the  node with the highest entailment score in the next layer. 
Second, evaluating the entailment scores of all possible parent-child pairs is expensive, so we retrieve the $3$ nearest neighbors of the node A (other than its matched node B) according to the \texttt{all-mpnet-base-v2} embeddings. If the entailment score between the retrieved node and the parent of A is above $\tau=0.7$, we add an arrow between them as shown in \Cref{fig:workflow} step~4. The very similar prompts are more likely to create similar parents and entail multiple parents, so another benefit of the DAG structure is to let the user easily visualize the groups of very similar prompts. 








\subsection{Bottom-up Construction}

After all the common prompts are extracted, the common prompts become the next level's prompt nodes. Next, we repeat matching, generation, and optional assignment until a single root remains (\Cref{fig:workflow}, Step~5). 
The result is a deep, fine-grained hierarchy whose every internal node is a prompt entailed by the entire sub-DAG below it.

Our bottom-up and layer-by-layer construction has several advantages. First, the common prompts are usually much shorter than the original user prompts, so building the top layers in our deep tree is faster and cheaper. Second, generating the specific common prompt is challenging. The LLM is often forced to output very general common prompts when it must generate a common prompt from more than two prompts, or if one of the prompts is not similar to the rest. Our method reduces the difficulty of generating very specific common prompt by ensuring that each common prompt is generated by two similar prompts.




%% file: contents/experiments.tex

\section{Experiments}


 





To facilitate the future development of prompt understanding approaches, we first propose several intrinsic evaluation metrics and conduct ablation studies in \Cref{sec:metrics}-\ref{sec:human_support}. Next, we evaluate how well \methodname helps another LLM as a human proxy to write the report of LLMs' strengths and weaknesses in \Cref{sec:agent_eval}.

\subsection{Intrinsic Evaluation Metrics}
\label{sec:metrics}

Our intrinsic metrics should allow us to compare the hierarchical categorization methods with different structures and the different specificity of their node descriptions. Different structures have different numbers of levels, so we evaluate the entailment scores from two randomly selected descendant leaf nodes rather than from its immediate children\footnote{The entailment scores between the parent and children tend to be higher in a high level because finding the common prompts between two short and general prompts is generally easier. 
} to measure how much the leaf user prompts actually belong to the cluster. 

Besides the average of two entailment scores called Entail$_{avg}$, we also compute their minimum Entail$_{min}$ to measure how likely the method only summarizes one of the child nodes. The evaluation directly tests the faithfulness of each cluster description to their leaf members and the transitivity of the entailment relation in the hierarchy. Taking \Cref{fig:workflow} as an example, we evaluate the node $F$ using the entailment score of $A\!\to\!F$ rather than $D\!\to\!F$ because the score of $D\!\to\!F$ tends to be higher than the score from $A\!\to\!D$. Assuming we want to compute the entailment score for node $F$ and two descendant leaf nodes are $A$ and $B$, Entail$^F_{avg}=(e_{A\!\to\!F} + e_{B\!\to\!F})/2$ and Entail$^F_{min}=\min(e_{A\!\to\!F}, e_{B\!\to\!F})$. 

Entailment scores alone are not enough because a method could output a very general cluster description (e.g., always output \textit{Solving a problem}) to get very high entailment scores. To solve the problem, we use the informativeness metrics from summarization and the specificity metric from \citet{chang2024fine}. That is, a good cluster description should be specific enough to contain important information of their descendant leaf nodes and have low entailment scores for non-descendant leaf nodes (i.e., the specific description should reduce the extra assignment links in step 4 in \Cref{fig:workflow}).
We use \texttt{GPT 5.4 Mini} with low reasoning effort to estimate both entailment score / faithfulness and informativeness. In short, we define the following metrics:
\begin{itemize}[itemsep=2pt, parsep=0pt, topsep=1pt,leftmargin=*]
    \item Entail$_{avg/min}$: The average/minimum of the entailment scores from two randomly selected descendant leaf nodes. The entailment score is the percentage of information in the upper node that is faithfully grounded in the lower node.
    \item Info$_{avg/min}$: The average/minimum percentage of the important information in the lower node that is retained by the upper node.
    \item Specificity: Entail$_{avg}$ - Entail$_{other}$, where Entail$_{other}$ is the average entailment score from the non-descendant leaf node that is most similar to one of the descendant leaf nodes (e.g., C in \Cref{fig:workflow}). 
    \item Len (length): The average number of words in non-leaf nodes.
\end{itemize}

\begin{table*}[t!]
    \centering
    \scalebox{0.8}{
    \begin{tabular}{l|cccc}
        \toprule
         Methods & Clustering & Leaf Main Task & Parent Generation   \\
         \midrule
        EvalTree & Recursive Kmeans & Capability & Capability   \\
        EvalTree + Our Prompts & Recursive Kmeans & Main Task & Common Prompt  \\
        EvalTree + Our Prompts + Org. & Recursive Kmeans & Original & Common Prompt \\
         \methodname~- Main - Eval  & Match & Original & Common Prompt \\
         \methodname~- Main  & Match + Assignment & Original & Common Prompt + Regen. \\
         \methodname  & Match + Main + Assignment & Original & Common Prompt + Regen. \\
         \midrule
         \methodname on Capability  & Match + Main + Assignment & Capability & Common Prompt + Regen. \\
         \methodname on Main Task  & Match + Main + Assignment & Main Task & Common Prompt + Regen. \\
         EvalTree + Long Prompt & Recursive Kmeans & Long Capability & Long Capability   \\
         EvalTree (MiniLM) & Recursive Kmeans & Capability & Capability   \\
         EvalTree + Long Prompt (MiniLM) & Recursive Kmeans & Long Capability & Long Capability  \\
         
         \bottomrule
    \end{tabular}
    }
    \caption{Differences of our ablation methods. The Assignment in clustering means Step 4. The Main means extracting the main task of the leaf nodes to improve similarity measurement in Step 1.}
    \label{tab:ablation}
\end{table*}

\subsection{Setup}

We conduct our experiments in WildBench~\citep{lin2025wildbench}, which contains more than a thousand complicated real user queries, and UltraFeedback~\citep{ultrafeedback}, which is one of the best LLM alignment datasets~\citep{ivison2024unpacking}. It is costly to evaluate on all prompts, so we randomly select a subset with $256$ samples for each dataset.  

Both \methodname and EvalTree use \texttt{all-mpnet-base-v2} to measure similarity and \texttt{Claude Sonnet 4.6} to generate the node description or main tasks. We set $k=2$ in recursive Kmeans in EvalTree to make it more comparable to our hierarchy structure.  During hierarchy construction, we predict the entailment scores using \texttt{GPT-4o Mini} rather than \texttt{GPT-5.4 mini} to avoid overfitting to our LLM evaluator.

\subsection{Baseline and Ablations}

The existing state-of-the-art prompt tree construction methods usually hierarchically cluster the user prompts and then ask LLMs to generate the short description of each cluster~\cite{tamkin2024clio,zeng2025evaltree}. EvalTree~\cite{zeng2025evaltree} is a representative open-source method among them, so we adopt it as our primary baseline and conduct an ablation study by gradually adding the proposed components to EvalTree~\cite{zeng2025evaltree}. 
We summarize the differences of baseline/ablations in \Cref{tab:ablation} and describe each method below:
\begin{itemize}[itemsep=2pt, parsep=0pt, topsep=1pt,leftmargin=*]
\item \textbf{EvalTree}: The method extracts the capability from the user prompts, clusters the prompts using recursive Kmeans, and generates a short capability description for each node.
\item \textbf{EvalTree + Our Prompts}: Replace EvalTree's prompt for capability extraction with our main task extraction and common prompt generation in \Cref{sec:common_prompt}. 
\item \textbf{EvalTree + Our Prompts + Org.}: After recursive Kmeans, directly extract the common prompts from the original user prompts rather than their summaries. 
\item \textbf{\methodname~- Main - Eval}: Instead of performing Kmeans, we interleave the matching and common prompt generation as described in \Cref{sec:matching}.
\item \textbf{\methodname~- Main}: Adding the steps that need another LLM for evaluation, including regeneration in \Cref{sec:common_prompt} and the optional assignment in \Cref{sec:optional}.
\item \textbf{\methodname}: Adding the similarity estimation using main tasks in \Cref{sec:matching} to get our full method.
\item \textbf{\methodname on Capability}: \methodname built on the same capability leaf compared to EvalTree.
\item \textbf{\methodname on Main Task}: \methodname built on the main task.
\item \textbf{EvalTree + Long Prompt}: Instruct the LLM to generate a longer capability description for each non-leaf node.
\item \textbf{EvalTree (MiniLM)}: To make sure the low performance of EvalTree does not come from an incompatible embedding model, we use \texttt{all-MiniLM-L6-v2} to replace \texttt{all-mpnet-base-v2} in EvalTree.
\item \textbf{EvalTree + Long Prompt (MiniLM)}: EvalTree + Long Prompt using the MiniLM embeddings.

\end{itemize}

\begin{table*}[t!]
    \centering
    \resizebox{\linewidth}{!}{
    \begin{tabular}{llrrrrrr}
        \toprule
        Data & Methods & Entail$_{avg}$ & Entail$_{min}$ & Info$_{avg}$ & Info$_{min}$ & Specificity & Len (words) \\
        \midrule
        \multirow{11}{*}{WB256} & EvalTree & 36.3$_{\pm 1.7}$ & 22.6$_{\pm 1.6}$ & 22.9$_{\pm 1.0}$ & 16.3$_{\pm 0.9}$ & 13.0$_{\pm 2.2}$ & 5.7$_{\pm 0.1}$ \\
        & EvalTree + Our Prompts & 47.1$_{\pm 1.6}$ & 32.9$_{\pm 1.8}$ & 36.3$_{\pm 1.4}$ & 26.3$_{\pm 1.4}$ & 25.4$_{\pm 2.1}$ & 15.8$_{\pm 0.3}$ \\
        & EvalTree + Our Prompts + Org. & 50.4$_{\pm 1.7}$ & 35.5$_{\pm 1.9}$ & 41.0$_{\pm 1.6}$ & 31.6$_{\pm 1.6}$ & 23.8$_{\pm 2.2}$ & 21.4$_{\pm 0.9}$ \\
        & \methodname~- Main - Eval & 53.5$_{\pm 1.7}$ & 38.8$_{\pm 1.9}$ & 44.2$_{\pm 1.5}$ & 34.9$_{\pm 1.6}$ & 26.4$_{\pm 2.4}$ & 21.3$_{\pm 1.0}$ \\
        & \methodname~- Main & 55.4$_{\pm 1.7}$ & 40.7$_{\pm 2.0}$ & 45.6$_{\pm 1.6}$ & 36.0$_{\pm 1.7}$ & 29.3$_{\pm 2.3}$ & 21.1$_{\pm 0.9}$ \\
        & \methodname & \textbf{55.9}$_{\pm 1.7}$ & \textbf{41.7}$_{\pm 1.9}$ & \textbf{49.1}$_{\pm 1.5}$ & \textbf{39.1}$_{\pm 1.7}$ & \textbf{32.5}$_{\pm 2.3}$ & 21.6$_{\pm 1.0}$ \\
        \cmidrule{2-8}
        & \methodname on Capability & 50.8$_{\pm 1.7}$ & 35.7$_{\pm 1.8}$ & 37.8$_{\pm 1.3}$ & 27.9$_{\pm 1.2}$ & 24.6$_{\pm 2.1}$ & 12.3$_{\pm 0.2}$ \\
        & \methodname on Main Task & 50.4$_{\pm 1.6}$ & 35.7$_{\pm 1.8}$ & 37.9$_{\pm 1.4}$ & 28.6$_{\pm 1.4}$ & 27.8$_{\pm 2.3}$ & 15.3$_{\pm 0.3}$ \\
        & EvalTree + Long Prompt &  29.9$_{\pm 1.4}$ & 17.6$_{\pm 1.3}$ & 41.7$_{\pm 1.4}$ & 29.5$_{\pm 1.5}$ & 17.3$_{\pm 1.7}$ & 19.4$_{\pm 0.1}$ \\
        & EvalTree (MiniLM) &  39.7$_{\pm 1.9}$ & 24.8$_{\pm 1.9}$ & 21.2$_{\pm 1.0}$ & 14.0$_{\pm 0.8}$ & 17.7$_{\pm 2.2}$ & 5.5$_{\pm 0.1}$ \\
        & EvalTree + Long Prompt (MiniLM) & 30.6$_{\pm 1.5}$ & 17.5$_{\pm 1.3}$ & 37.4$_{\pm 1.4}$ & 26.8$_{\pm 1.5}$ & 18.0$_{\pm 1.6}$ & 19.5$_{\pm 0.1}$  \\

        
        \midrule
        \multirow{11}{*}{UF256} & EvalTree & 36.2$_{\pm 1.9}$ & 23.1$_{\pm 1.9}$ & 26.1$_{\pm 1.4}$ & 18.5$_{\pm 1.2}$ & 14.4$_{\pm 2.4}$ & 5.4$_{\pm 0.1}$ \\
        & EvalTree + Our Prompts & 41.5$_{\pm 1.7}$ & 26.9$_{\pm 1.7}$ & 31.6$_{\pm 1.4}$ & 22.0$_{\pm 1.3}$ & 20.0$_{\pm 2.1}$ & 12.5$_{\pm 0.3}$ \\
        & EvalTree + Our Prompts + Org. & 52.0$_{\pm 1.8}$ & 37.6$_{\pm 2.0}$ & 36.8$_{\pm 1.5}$ & 27.3$_{\pm 1.5}$ & 24.0$_{\pm 2.4}$ & 14.5$_{\pm 0.4}$ \\
        & \methodname~- Main - Eval & \textbf{55.6}$_{\pm 1.7}$ & 41.1$_{\pm 1.9}$ & 41.1$_{\pm 1.5}$ & 31.8$_{\pm 1.6}$ & 28.4$_{\pm 2.4}$ & 14.3$_{\pm 0.3}$ \\
        & \methodname~- Main & 54.4$_{\pm 1.7}$ & 40.3$_{\pm 2.0}$ & 43.7$_{\pm 1.6}$ & 34.5$_{\pm 1.6}$ & 31.1$_{\pm 2.5}$ & 14.5$_{\pm 0.3}$ \\
        & \methodname & 55.3$_{\pm 1.7}$ & \textbf{41.8}$_{\pm 2.0}$ & \textbf{44.5}$_{\pm 1.6}$ & \textbf{35.5}$_{\pm 1.7}$ & \textbf{31.4}$_{\pm 2.4}$ & 14.0$_{\pm 0.4}$  \\
        \cmidrule{2-8}
        & \methodname on Capability & 49.2$_{\pm 1.7}$ & 32.9$_{\pm 1.9}$ & 36.9$_{\pm 1.4}$ & 26.4$_{\pm 1.3}$  & 22.1$_{\pm 2.4}$ & 10.8$_{\pm 0.2}$ \\
        & \methodname on Main Task & 46.9$_{\pm 1.8}$ & 33.0$_{\pm 2.0}$ & 33.8$_{\pm 1.5}$ & 25.7$_{\pm 1.4}$ &  25.7$_{\pm 2.2}$ & 12.1$_{\pm 0.3}$ \\
        & EvalTree + Long Prompt  & 30.2$_{\pm 1.6}$ & 19.9$_{\pm 1.5}$ & 42.8$_{\pm 1.7}$ & 31.9$_{\pm 1.7}$ & 14.9$_{\pm 1.9}$ & 18.9$_{\pm 0.1}$ \\
        & EvalTree (MiniLM) &  37.6$_{\pm 1.9}$ & 24.5$_{\pm 1.9}$ & 25.5$_{\pm 1.3}$ & 18.0$_{\pm 1.2}$ & 15.9$_{\pm 2.4}$ & 5.6$_{\pm 0.1}$ \\
        & EvalTree + Long Prompt (MiniLM) & 31.6$_{\pm 1.6}$ & 19.3$_{\pm 1.5}$ & 40.7$_{\pm 1.6}$ & 28.8$_{\pm 1.7}$ & 18.2$_{\pm 1.8}$ & 18.9$_{\pm 0.1}$ \\
        \bottomrule

    \end{tabular}
    }
    \caption{Comparison of EvalTree, \methodname, and its ablations on 256 randomly sampled prompts from WildBench (WB) and UltraFeedback (UF). Values are mean $\pm$ standard error. Entailment and informativeness are percentages, Specificity is in percentage points, and Len is in words. The best score for each metric within each dataset is highlighted.}
    \label{tab:main_results}
\end{table*}

\begin{table}[t!]
\parbox{.48\linewidth}{
    \centering
    \scalebox{0.8}{
    \begin{tabular}{l|cccc}
        \toprule
         & \multicolumn{2}{c}{In-Cluster}  & \multicolumn{2}{c}{Out-of-Cluster} \\
         & Entail$_{avg}$ & Info$_{avg}$ & Entail & Info \\
         \midrule
         Human & 66.4$_{\pm 3.3}$ & 43.4$_{\pm 2.5}$ & 58.9$_{\pm 2.9}$ & 39.2$_{\pm 2.5}$ \\
         LLM & 65.7$_{\pm 5.4}$ & 37.3$_{\pm 3.9}$ & 59.8$_{\pm 5.1}$ & 40.6$_{\pm 3.5}$ \\
         \bottomrule
    \end{tabular}
    }
    \caption{Comparison between the average scores from humans and from our LLM judge. The out-of-cluster scores come from the non-descendant node that is most similar to one of the cluster nodes.}
    \label{tab:human_avg}
}
\hfill
\parbox{.48\linewidth}{
    \centering
    \scalebox{0.8}{
    \begin{tabular}{l|cccc}
        \toprule
        & \multicolumn{2}{c}{Human}  & \multicolumn{2}{c}{Human Clean} \\
         & Entail & Info & Entail & Info \\
         \midrule
         r(H,H) & 0.12 & 0.16 & \textbf{0.31} & 0.55 \\
         r(H, LLM) & 0.29 & 0.42 & 0.21 & 0.59 \\
         r(H avg, LLM) & \textbf{0.43} & \textbf{0.59} & \textbf{0.31} & \textbf{0.69} \\
         \bottomrule
    \end{tabular}
    }
    \caption{The average Pearson correlations between two workers, r(H,H), between one worker and our LLM judge score, r(H, LLM), and between the average score of all workers and the LLM score, r(H avg, LLM). }
    \label{tab:human_corr}
}
\end{table}

\subsection{Results}
\label{sec:eval_results}

\Cref{tab:main_results} shows that \methodname performs drastically better than \textbf{EvalTree} in all metrics.
It even doubles the specificity and informativeness metrics of \textbf{EvalTree} in WildBench. In both WildBench and UltraFeedback, \textbf{\methodname} improves the Entail$_{avg}$ of \textbf{EvalTree} by almost $20$. The performance of \textbf{EvalTree + Long Prompt} demonstrates that simply increasing the summary lengths of EvalTree by naive prompt engineering could only increase the informativeness at the cost of the faithfulness/entailment score. The similar performances of \textbf{EvalTree (MiniLM)} and \textbf{EvalTree} show that their worse performances do not come from their sensitivity to the clustering embedding space.

The ablation methods suggest where the improvements come from. The first big gap between \textbf{EvalTree} and \textbf{EvalTree + Our Prompts + Org.} indicates that generating the longer and more specific descriptions using our prompt improves all the metrics, and starting from the original user prompts usually makes descriptions more specific (i.e., increasing the informativeness). The much better performances of \textbf{\methodname} compared to \textbf{\methodname on Capability} and \textbf{\methodname on Main Task} also support this observation.  

\textbf{\methodname on Main Task} is better than \textbf{EvalTree + Our Prompts}.
Furthermore, \textbf{\methodname~- Main - Eval} often wins more over \textbf{EvalTree + Our Prompts + Org.} in almost all metrics.  These demonstrate that our construction method, which alternatively matches and generates the common prompts, creates not only a more balanced hierarchy but also better cluster descriptions, which is especially important for the longer prompts. 

The similar or slightly better performances of \textbf{\methodname~- Main} compared with \textbf{\methodname~- Main - Eval} demonstrate that our main improvements do not come from the optional assignment and the evaluation-based regeneration. 










\subsection{Human Evaluation}
\label{sec:human_support}

We conduct a human study using Amazon Mechanical Turk (MTurk) to verify the entailment and informativeness scores in WildBench from the LLM as a judge. Each task asks the human worker to score the entailment/faithfulness and informativeness of A, B, and C to their common prompt parent in \Cref{fig:workflow} that is generated by \methodname, where A and B are in-cluster nodes and C is the out-of-cluster node. Thirty tasks with different common prompts are launched, and each task is done by at least $3$ workers. We control the quality first using the master worker labels and manually rejecting the obviously unqualified responses. Next, we use Claude Code to identify low-quality workers. 

In \Cref{tab:human_avg}, we compare the average scores from humans and the \texttt{GPT-5.4 mini} LLM judge. The LLM provides very similar average scores for other metrics except for underestimating in-cluster informativeness. The left two columns of \Cref{tab:human_corr} show that the correlation between the LLM and humans, r(H, LLM), is higher than inter-annotator agreement, r(H, H). Furthermore, the correlation would be even higher if we reduce the human personal biases by averaging their scores, r(H avg, LLM). If we remove the borderline annotations, the inter-annotator agreements increase but their informativeness is still lower than the r(H avg, LLM). The result validates the effectiveness of our LLM judge.








\begin{figure}[t!]
  \centering
  \begin{minipage}{0.45\linewidth}
    \centering
    \includegraphics[width=\linewidth]{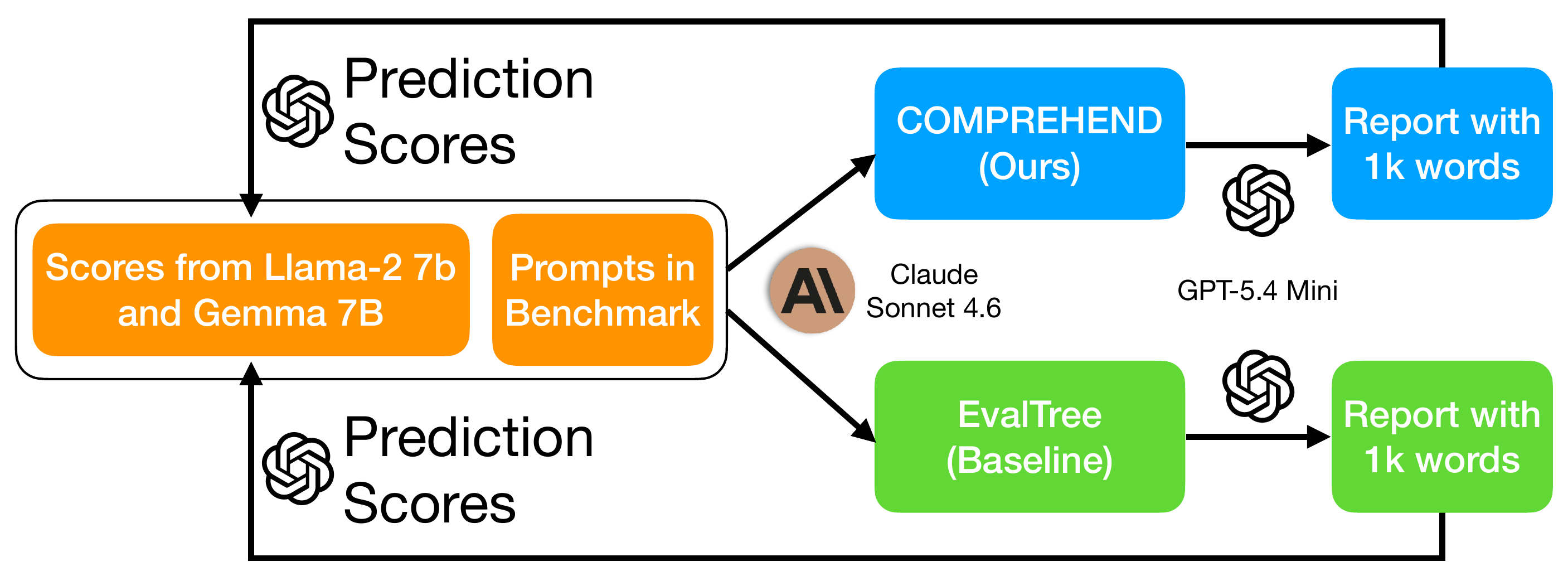}
    \caption{Our evaluation workflow using an LLM agent.}
    \label{fig:agentic_eval}
  \end{minipage}\hfill
  \begin{minipage}{0.48\linewidth}
    \centering
    \scalebox{0.8}{%
      \begin{tabular}{c|cc}
        \toprule
        & RMSE ($\downarrow$) & Pearson (\%) \\
        \midrule
        \multicolumn{3}{c}{Llama-2 7B} \\
        \midrule
        EvalTree & 1.83$_{\pm 0.17}$ & 36.0 \\
        \methodname & \textbf{1.74}$_{\pm 0.17}$ & \textbf{45.0} \\
        \midrule
        \multicolumn{3}{c}{Gemma 7B} \\
        \midrule
        EvalTree & 2.26$_{\pm 0.16}$ & 18.9 \\
        \methodname & \textbf{2.18}$_{\pm 0.17}$ & \textbf{20.2} \\
        \bottomrule
      \end{tabular}%
    }
    \captionof{table}{The comparison using an LLM agent. Lower RMSE and higher Pearson correlation mean that the report derived from the analysis tool is better at predicting the original LLM score.}
    \label{tab:agent}
  \end{minipage}
\end{figure}

\subsection{Benchmark Understanding by an LLM}
\label{sec:agent_eval}

One main goal of \methodname is to let humans discover the strengths and weaknesses of LLMs. In this section, we use an LLM as an agent to compare \methodname and EvalTree for analyzing the performance of \texttt{Gemma 7b It} or \texttt{Llama-2 7b Chat} in a WildBench subset with 64 prompts. 
As shown in \Cref{fig:agentic_eval}, we ask \texttt{GPT-5.4 Mini} to write a report with around $1000$ words 
from a json file of \methodname or EvalTree that stores $63$ cluster descriptions, average performance of \texttt{Gemma} or \texttt{Llama-2}, and the edges between the nodes (see our visualizations of a 1024 prompts in \Cref{fig:ours_1024_compare}). 

Next, we evaluate the quality of the reports by predicting the raw scores of 64 user prompts. A faithful and informative hierarchy should help the \texttt{GPT-5.4 Mini} to write a report that can reconstruct the original scores well and thus have a low RMSE (Root Mean Square Error) and high Pearson correlation coefficient in its score prediction. The results in \Cref{tab:agent} show that \textbf{\methodname} gains large improvements in the Pearson correlation over \textbf{EvalTree} in \texttt{Llama-2}.





%% file: contents/conclusion_limitation.tex
\section{Conclusion}

Prior work such as EvalTree hierarchically clusters the prompts and generates the short and general description of the clusters to facilitate humans' understanding of LLMs' performances on user prompts. Each description could be viewed as a summary of all the user prompts in the cluster, so we propose to evaluate the description using entailment/faithfulness and informativeness. Our experiments demonstrate that both instructing an LLM to generate more specific common prompt and iteratively conducting matching and description generation from bottom to top not only lead to a more balanced hierarchy but also drastically improve the faithfulness and informativeness of the cluster descriptions, which helps an LLM agent to generate a better performance report in a benchmark.




%% file: appendix/more_results.tex

\begin{figure*}[t!]
\centering
\begin{minipage}{.48\textwidth}
  \centering
  \includegraphics[width=.99\linewidth]{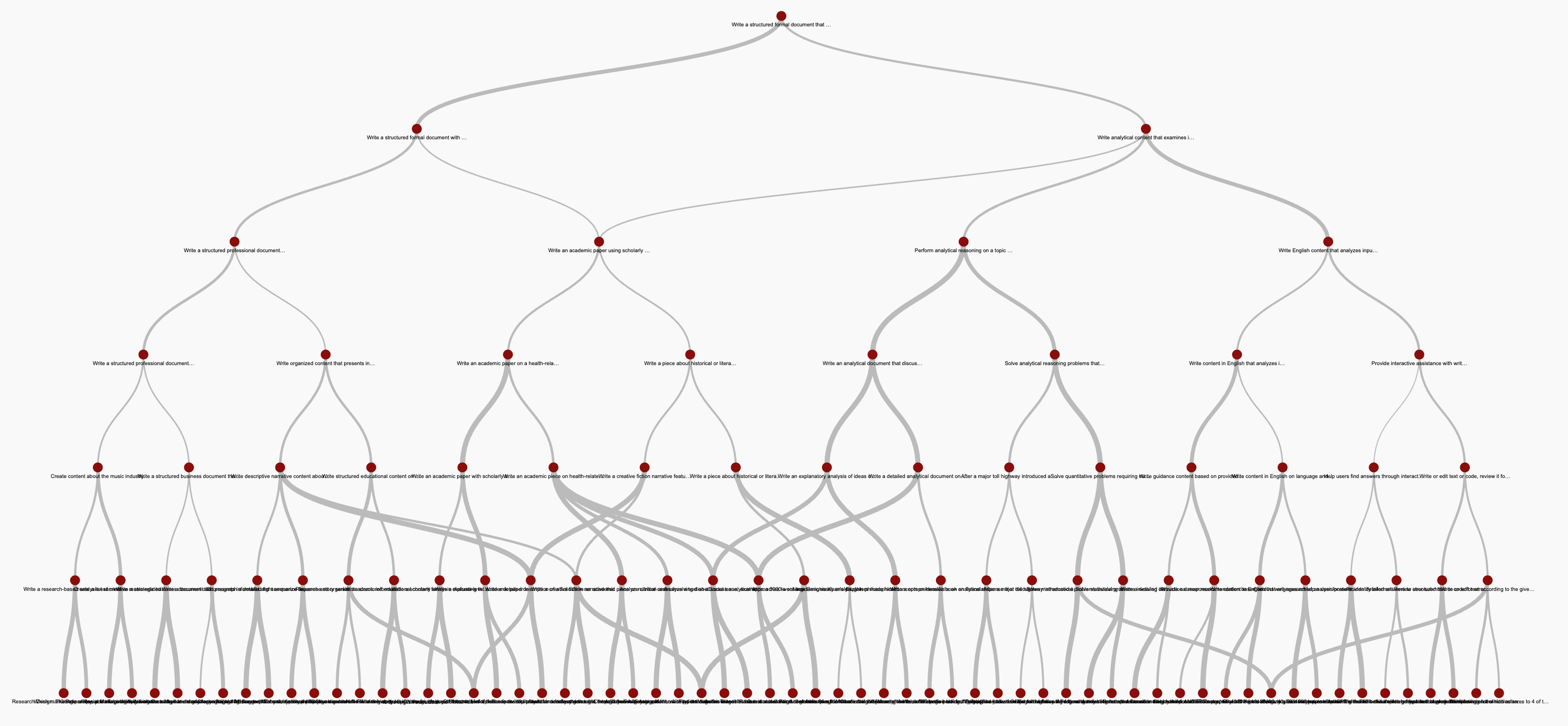}
  \captionof{figure}{\methodname for the dissimilar subset (i.e., 64 randomly sampled prompts in WildBench).}
  \label{fig:our_tree}
\end{minipage}%
\hspace{0.03\textwidth} 
\begin{minipage}{.48\textwidth}
  \centering
  \includegraphics[width=.99\linewidth]{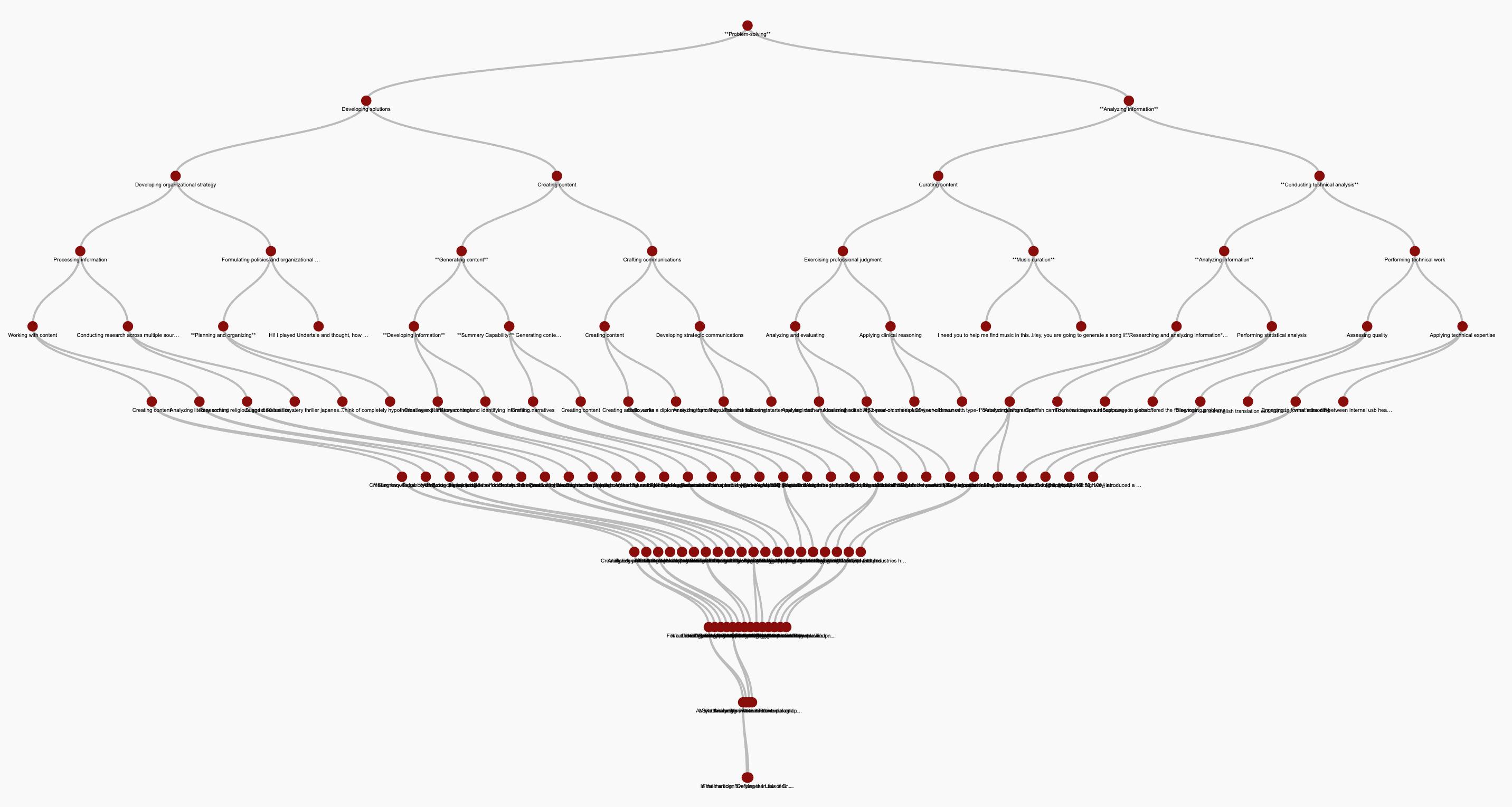}
  \captionof{figure}{EvalTree for the dissimilar subset.}
  \label{fig:eval_tree}
\end{minipage}
\end{figure*}

\begin{figure}[t]
  \includegraphics[width=.99\linewidth]{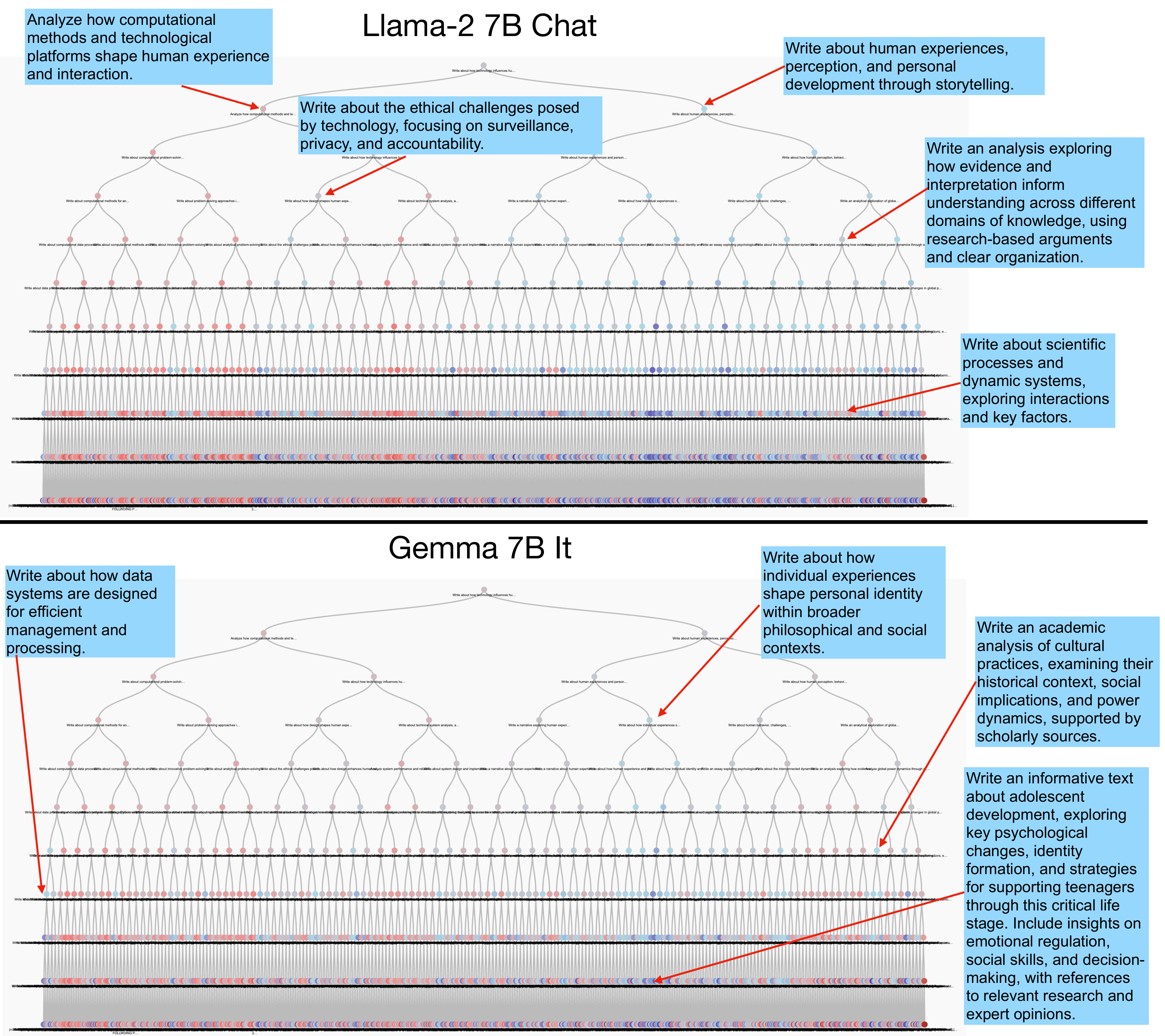}
  \captionof{figure}{Example of analyzing the strengths and weaknesses of LLMs. Blue represents a high score while red means lower performance.}
  \label{fig:ours_1024_compare}
\end{figure}

\begin{figure}[t]
  \includegraphics[width=.99\linewidth]{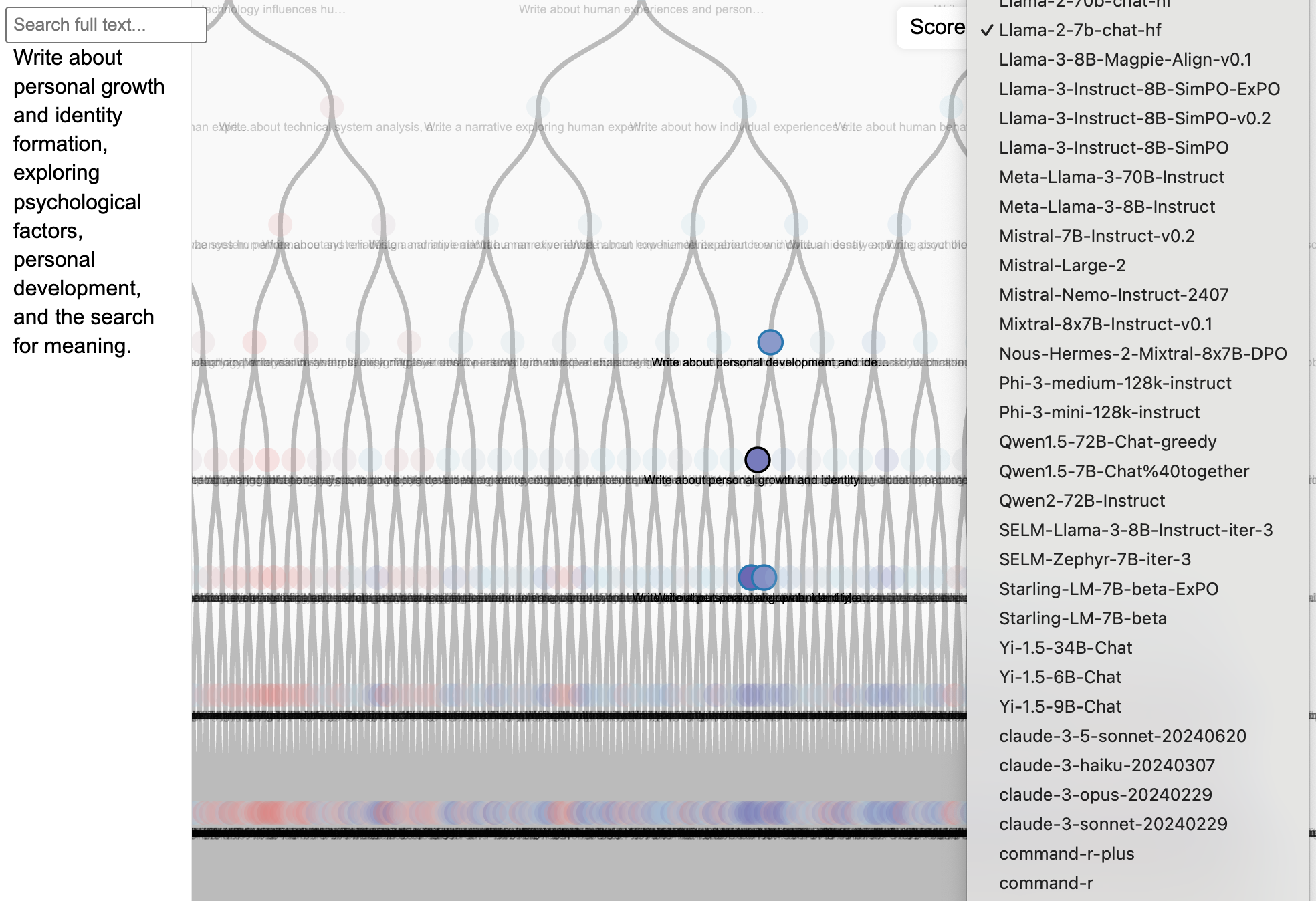}
  \captionof{figure}{Our visualization tool allows users to easily analyze the performance of different LLMs on different prompts in the entailment hierarchy.}
  \label{fig:ours_1024_zoom}
\end{figure}

\section{Visualization}
\label{sec:visualization}

In \Cref{fig:compare_first}, we compare cluster descriptions of \methodname and EvalTree qualitatively. Here, we visualize their structures in \Cref{fig:our_tree} and \Cref{fig:eval_tree}, where a thicker edge indicates a larger entailment score. The construction of \methodname makes the same-level nodes to be the clusters of similar sizes, while EvalTree could have small clusters at a high level. Furthermore, some regions of \methodname have more assignment links, which suggest that prompts in the region are more similar to each other as described in \Cref{sec:optional}.

We also implement a simple weakness analysis tool and show its screenshot in \Cref{fig:ours_1024_zoom}. The user can choose the LLMs they want to analyze and its average score in every cluster would be visualized using color (Dark blue is high and dark red is low). The sidebar will show the description of the clicked node and their neighbors will be highlighted. In \Cref{fig:ours_1024_compare}, we compare the \texttt{Gemma} and \texttt{Llama-2} on the $1024$ user prompts in WildBench. We can see that high-performance blue nodes are clustered together and \texttt{Llama 2} has better creative writing and worse reasoning ability compared to \texttt{Gemma}. You can download our visualization from the ./demo\_trees folder in our GitHub repository.

\begin{table*}[t!]
\centering
\resizebox{\linewidth}{!}{\begin{tabular}{lllrrrrrr}
\toprule
Data & Generation model & Method & Entail$_{avg}$ & Entail$_{min}$ & Info$_{avg}$ & Info$_{min}$ & Specificity & Len (words) \\
\midrule
\multirow{9}{*}{WB256} & \multirow{3}{*}{Claude Sonnet 4.6} & EvalTree & 36.3$_{\pm1.7}$ & 22.6$_{\pm1.6}$ & 22.9$_{\pm1.0}$ & 16.3$_{\pm0.9}$ & 13.0$_{\pm2.2}$ & 5.69$_{\pm0.13}$ \\
&& EvalTree (MiniLM) & 39.7$_{\pm 1.9}$ & 24.8$_{\pm 1.9}$ & 21.2$_{\pm 1.0}$ & 14.0$_{\pm 0.8}$ & 17.7$_{\pm 2.2}$ & 5.46$_{\pm 0.12}$ \\
& & \methodname & 55.9$_{\pm1.7}$ & 41.7$_{\pm1.9}$ & \textbf{49.1}$_{\pm1.5}$ & \textbf{39.1}$_{\pm1.7}$ & 32.5$_{\pm2.3}$ & 21.64$_{\pm1.02}$ \\
\cmidrule{2-9}
& \multirow{3}{*}{GLM 5.2} & EvalTree & 38.3$_{\pm1.9}$ & 23.8$_{\pm1.9}$ & 16.1$_{\pm0.8}$ & 11.2$_{\pm0.6}$ & 13.8$_{\pm2.3}$ & 2.96$_{\pm0.04}$ \\
&& EvalTree (MiniLM) & 36.9$_{\pm 2.0}$ & 22.5$_{\pm 1.9}$ & 14.5$_{\pm 0.7}$ & 10.5$_{\pm 0.5}$ & 7.7$_{\pm 2.6}$  & 2.96$_{\pm 0.04}$   \\
& & \methodname & \textbf{60.5}$_{\pm1.8}$ & \textbf{47.9}$_{\pm2.0}$ & 40.7$_{\pm1.5}$ & 32.2$_{\pm1.5}$ & \textbf{33.5}$_{\pm2.6}$ & 15.33$_{\pm1.19}$ \\
\cmidrule{2-9}
& \multirow{3}{*}{DeepSeek V4 Flash} & EvalTree & 35.6$_{\pm1.9}$ & 21.7$_{\pm1.8}$ & 15.0$_{\pm0.8}$ & 10.0$_{\pm0.5}$ & 10.7$_{\pm2.4}$ & 2.77$_{\pm0.04}$ \\
&& EvalTree (MiniLM) & 39.1$_{\pm2.0}$  & 24.2$_{\pm1.9}$ & 13.0$_{\pm0.6}$ & 8.6$_{\pm0.4}$ & 12.5$_{\pm2.4}$ & 2.85$_{\pm0.04}$ \\
& & \methodname & 52.8$_{\pm1.7}$ & 38.3$_{\pm1.9}$ & 36.5$_{\pm1.5}$ & 29.1$_{\pm1.5}$ & 30.9$_{\pm2.4}$ & 17.15$_{\pm0.84}$ \\

\midrule
\multirow{9}{*}{UF256} & \multirow{3}{*}{Claude Sonnet 4.6} & EvalTree & 36.2$_{\pm1.9}$ & 23.1$_{\pm1.9}$ & 26.1$_{\pm1.4}$ & 18.5$_{\pm1.2}$ & 14.4$_{\pm2.4}$ & 5.41$_{\pm0.12}$ \\
&& EvalTree (MiniLM) & 37.6$_{\pm 1.9}$ & 24.5$_{\pm 1.9}$ & 25.5$_{\pm 1.3}$ & 18.0$_{\pm 1.2}$ & 15.9$_{\pm 2.4}$ & 5.59$_{\pm 0.10}$ \\
& & \methodname & \textbf{55.3}$_{\pm1.7}$ & \textbf{41.8}$_{\pm2.0}$ & \textbf{44.5}$_{\pm1.6}$ & \textbf{35.5}$_{\pm1.7}$ & \textbf{31.4}$_{\pm2.4}$ & 13.97$_{\pm0.37}$ \\
\cmidrule{2-9}
& \multirow{3}{*}{GLM 5.2} & EvalTree & 42.1$_{\pm2.0}$ & 26.7$_{\pm2.0}$ & 19.5$_{\pm1.1}$ & 13.5$_{\pm0.9}$ & 11.1$_{\pm2.7}$ & 2.99$_{\pm0.05}$ \\
&& EvalTree (MiniLM) & 41.4$_{\pm 2.1}$ & 27.1$_{\pm 2.1}$ & 16.6$_{\pm 1.0}$ & 11.3$_{\pm 0.8}$ & 11.3$_{\pm 2.6}$  &2.78$_{\pm 0.04}$ \\
& & \methodname & 51.2$_{\pm1.8}$ & 38.1$_{\pm2.0}$ & 35.3$_{\pm1.6}$ & 27.3$_{\pm1.5}$ & 28.3$_{\pm2.4}$ & 10.13$_{\pm0.35}$ \\
\cmidrule{2-9}
& \multirow{3}{*}{DeepSeek V4 Flash} & EvalTree & 41.6$_{\pm2.0}$ & 27.1$_{\pm2.0}$ & 18.7$_{\pm1.1}$ & 13.2$_{\pm1.0}$ & 10.4$_{\pm2.8}$ & 2.69$_{\pm0.04}$  \\
&& EvalTree (MiniLM) &   41.4$_{\pm 2.1}$ & 26.8$_{\pm 2.2}$ & 16.8$_{\pm 0.9}$ & 11.3$_{\pm 0.7}$ & 11.5$_{\pm 2.8}$  & 2.66$_{\pm 0.04}$  \\
& & \methodname & 53.6$_{\pm1.8}$ & 38.8$_{\pm2.1}$ & 39.8$_{\pm1.7}$ & 30.9$_{\pm1.7}$ & 28.4$_{\pm2.5}$ & 13.48$_{\pm0.45}$ \\

\bottomrule
\end{tabular}}
\caption{Cross-model comparison on 256 randomly sampled prompts from WildBench (WB256) and UltraFeedback (UF256). Values are mean $\pm$ standard error. Entailment, informativeness, and Specificity are percentages, and Len is the number of words in each summary. The best score within each dataset is highlighted.}
\label{tab:cross_model_256}
\end{table*}




\section{Sensitivity to Summary-Generation LLMs}

Although \texttt{Claude Sonnet 4.6} performs well, its API could be expensive for building a large entailment hierarchy.
We compare \texttt{GLM 5.2} and \texttt{DeepSeek V4 Flash} with \texttt{Claude Sonnet 4.6} on WB256 and UF256 in \Cref{tab:cross_model_256}. All comparisons use $k=2$ for EvalTree, and \texttt{GPT-5.4-mini} evaluation. Within every comparison of models and datasets, \methodname has higher mean entailment, informativeness, and Specificity than EvalTree. This shows that our conclusion holds across LLMs and proprietary models such as \texttt{Claude Sonnet 4.6} often outperform strong open models by a little.

\section{Sensitivity to EvalTree Cluster Size}
\label{sec:k_sensitivity}

To examine whether the comparison depends on EvalTree's cluster-size setting, we evaluate $k\in\{2,4,6,8\}$ on the same fixed random sample of 64 WildBench prompts (WB64). We use MPNet embeddings, \texttt{Claude Sonnet 4.6} generation, and \texttt{GPT-5.4 Mini} evaluation with two sampled descendant leaves per internal node.
This experiment is separate from the WB256/UF256 evaluations.

\begin{table*}[!htbp]
\centering
\resizebox{\linewidth}{!}{\begin{tabular}{lrrrrrrr}
\toprule
Method & Parents & Entail$_{avg}$ & Entail$_{min}$ & Info$_{avg}$ & Info$_{min}$ & Specificity & Len (words) \\
\midrule
EvalTree ($k=2$) & 63 & 31.2$_{\pm3.5}$ & 19.9$_{\pm3.2}$ & \textbf{20.6}$_{\pm2.0}$ & 13.2$_{\pm1.4}$ & 25.5$_{\pm3.7}$ & 6.5$_{\pm0.3}$ \\
EvalTree ($k=4$) & 30 & \textbf{32.1}$_{\pm5.5}$ & \textbf{20.2}$_{\pm5.0}$ & 19.9$_{\pm2.7}$ & \textbf{14.0}$_{\pm2.1}$ & \textbf{27.4}$_{\pm5.2}$ & 7.1$_{\pm0.4}$ \\
EvalTree ($k=6$) & 20 & 28.8$_{\pm5.9}$ & 17.6$_{\pm5.2}$ & 18.1$_{\pm2.2}$ & 12.7$_{\pm1.9}$ & 24.3$_{\pm5.7}$ & 7.8$_{\pm0.5}$ \\
EvalTree ($k=8$) & 13 & 24.5$_{\pm8.9}$ & 19.9$_{\pm8.8}$ & 13.7$_{\pm1.4}$ & 10.9$_{\pm1.7}$ & 22.0$_{\pm9.4}$ & 7.8$_{\pm0.7}$ \\
\bottomrule
\end{tabular}}
\caption{EvalTree cluster-size sensitivity on WB64. Values are mean $\pm$ descriptive standard error across evaluated non-leaf nodes. Entailment, informativeness, and Specificity are percentages.
Bold marks the highest mean in each score column.
}
\label{tab:k_sensitivity}
\end{table*}

To make the comparison fair, we set $k=2$ for EvalTree. As shown in \Cref{tab:k_sensitivity}, 
EvalTree's average entailment is highest at $k=4$, while its average informativeness is highest at $k=2$. The similar performances suggest that the observed entailment and informativeness advantages is unlikely to come from the choice of the $k$ value.

%% file: appendix/limitations.tex
\section{Limitations}

In theory, \methodname could derive an ultra-fine-grained understanding of any corpus after choosing the proper LLM, similarity model, and the LLM prompts. For example, when the corpus is the execution log of LLM agents, we can use \methodname to understand when the agents fail.
Furthermore, \methodname might be used to help LLM routing~\citep{okamoto2026trust} and select fine-tuning examples~\citep{cheng2024autodetect,zeng2025evaltree,zhang2026skill}. However, we currently only evaluate \methodname on WildBench and UltraFeedback.



The common prompt generation task is difficult. The LLMs still sometimes output the union of the two input prompts or the summary of just one of the two prompts. To fix the problem, we might need some more human examples to further fine-tune the LLMs.


Finally, some prompts might be more specific than the others~\citep{bhuiya2026prompt2box} and our method places all prompts at the bottom level. 
Besides, we always output the full hierarchy regardless of the similarity distribution of the prompts in the benchmark. 
If there are many very similar prompts in a benchmark, some of their parent nodes might be redundant. 

%% file: appendix/method_details.tex
\section{Method Details}
\label{sec:method_details}

We present our main task extraction prompt for step 1 in Template \ref{prompt:main_task} and the common prompt generation prompt for step 3 in Template \ref{prompt:common_prompt}. Step 4 is optional because we can skip this step to save cost by subsampling the dataset to be the closest $2^n$ when the dataset size is large and estimating the accurate entailment score could be expensive. 
Finally, as shown in \Cref{fig:ours_1024_zoom}, using our visualization tool, the users can choose the LLM at the top right corner to visualize their response scores, and search text at the top left corner. When the user clicks one node, its description is shown on the left and its parent and children will be highlighted.






\begin{myprop}
\label{prompt:main_task}
\begin{lstlisting}

Main Task Extraction (Batch Mode) 

You will be given a JSON object with key "texts" whose value is an array of texts.
For EACH text, extract the single most important main idea in ONE short sentence.

Output MUST be a JSON array of strings, same length and same order as the input texts.
Do NOT include any extra keys or commentary.

\end{lstlisting}
\end{myprop}

\begin{myprop}
\label{prompt:common_prompt}
\begin{lstlisting}

Common Prompt Generation

System message:
You are a professional editor who writes quality summary that generalizes two prompts A and B in as much specific detail as possible, while making sure that whatever is summarised is still true for both prompt A and B collectively (don't be too specific to one prompt, but while doing that also don't just union both prompts as a summary). No matter the content of prompts A & B, you must write a summary. DO NOT respond with messages that you can't or won't write a summary. That is simply not allowed, under any circumstances.

User message:
You will be given prompts A and B. Your task is to generate a prompt C where prompt C is implied by BOTH prompt A and prompt B. This means every detail in prompt C is directly mentioned in BOTH prompt A and B. Your prompt C should be as concise as possible with the following caveats: the more similar prompt A and B are too each other, the longer prompt C should be as it would include more details common to both prompt A and B. The more broad prompt A and B are, the more concise and general your prompt C should be. If your prompt A and B are so general, instead of summarizing prompt A and B, find a one sentence category that encapsulates the two prompts that includes the medium and the topic the user requests. Understand that you have to be as detailed as possible, while also not being too specific to just one prompt. Be detailed, but both children prompts should be entailed as much as possible by the summary you make (so high recall). Your summary should also be discriminative: always include the shared medium (writing/coding/analysis/etc.), the shared subject domain, and any key constraints common to both A and B — so that a prompt on a completely different topic could NOT plausibly imply your summary. Begin your prompt C with "(start)" and end it with "(end)."

1. (Dissimilar – Writing vs Poetry)
Prompt A:
"Write a detailed book review (800–1000 words) of George Orwell’s 1984. Focus on Orwell’s use of language, the historical context of the novel, and its relevance to modern surveillance culture. Please organize the review into an introduction, several analytical sections, and a concluding evaluation."
Prompt B:
"Compose a 12-stanza poem in free verse about a traveler crossing a desert, emphasizing themes of endurance, loneliness, and the harsh beauty of nature. Use vivid imagery and avoid rhyme schemes."
Prompt C:

(start) Write a structured piece of creative or analytical writing in English that explores thematic and human experiences using a specific literary form. (end)

2. (Dissimilar – Coding in Different Languages)
Prompt A:
"Can you implement a Python script that scrapes product data from an e-commerce website (like titles, prices, and availability) using BeautifulSoup? Please also save the data into a CSV file with properly labeled columns."
Prompt B:
"I’d like a C++ program that simulates a basic banking system, allowing users to create accounts, deposit money, withdraw money, and view balances. The program should be menu-driven and use object-oriented design."
Prompt C:

(start) Write a program that processes and manages structured data, implementing user-facing operations with organized output. (end)

3. (Moderately Similar – Writing)
Prompt A:
"Write a 2,000-word research paper about how social media influences political polarization in the United States. Discuss both positive and negative effects, provide examples from the past decade, and include at least five scholarly sources formatted in APA style."
Prompt B:
"Create a detailed argumentative essay about how modern technology (including social media, smartphones, and online forums) affects democracy. Focus on both risks and opportunities, use evidence from real-world examples, and cite at least three academic sources."
Prompt C:

(start) Write an essay on how social media affects politics and democracy, addressing both risks and benefits and including academic sources. (end)

4. (Moderately Similar – Coding)
Prompt A:
"Please create a Python script that takes a CSV file of sales transactions and generates summary statistics, including total revenue, average order value, and number of unique customers. Output the results to the terminal and also save them to a new CSV file."
Prompt B:
"Can you write a Python program that reads data from a JSON file containing customer purchases, calculates metrics like total sales and number of customers, and then produces a summary report saved to a text file?"
Prompt C:

(start) Write a Python script that reads purchase data, calculates total sales and customer counts, and outputs a summary report. (end)

5. (Very Similar – Writing)
Prompt A:
"Draft a 1,200-word persuasive essay arguing why renewable energy should replace fossil fuels as the dominant global energy source. The essay should include an introduction, three body sections (environmental benefits, economic advantages, and long-term sustainability), and a conclusion. Use real-world data and cite at least three credible sources."
Prompt B:
"Please write a well-structured essay, around 1,200 words, making the case for transitioning from fossil fuels to renewable energy. Discuss the environmental necessity, economic opportunities, and sustainable future benefits. Provide at least three reliable citations and organize the essay with intro, body, and conclusion."
Prompt C:

(start) Write a 1,200-word essay on why renewable energy should replace fossil fuels, discussing environmental benefits, economic advantages, and sustainability. Include at least three credible sources, with a structured introduction, body, and conclusion. (end)

6. (Very Similar – Coding)
Prompt A:
"Write a Python program that implements a REST API using Flask. The API should support basic CRUD operations for a task management app, including creating tasks, listing tasks, updating tasks, and deleting tasks. Store tasks in an in-memory dictionary for simplicity, and return JSON responses with appropriate HTTP status codes. Include comments and error handling."
Prompt B:
"I need a Python REST API built with Flask that can handle a simple to-do list. It should let users create tasks, retrieve all tasks, update them, and delete them. Use a dictionary to store tasks (no database needed), respond with JSON, and make sure to handle errors gracefully. Please comment the code clearly."
Prompt C:

(start) Write a Python REST API using Flask for a to-do app with CRUD operations, storing tasks in a dictionary, returning JSON responses, and including error handling and comments. (end)

Example 7
Prompt A (generalized)
Write about how different programming paradigms influence the way developers design and structure their code. Discuss the benefits and drawbacks of these approaches and how they affect maintainability and performance.
Prompt B (generalized)
Explain how programming paradigms shape software development. Highlight the trade-offs each paradigm brings and how they influence scalability and readability.
Prompt C (very broad)
(start) Write about how programming paradigms affect software design and development. (end)

Example 8
Prompt A (generalized)
Discuss the impact of artificial intelligence on fields traditionally driven by humans. Include opportunities, risks, and ethical implications.
Prompt B (generalized)
Write about the role of AI in human-centered domains, focusing on how it changes processes, raises ethical questions, and creates new forms of collaboration.
Prompt C (very broad)
(start) Write about the influence of AI on human-driven fields and its ethical implications. (end)

Example 9
Prompt A (generalized)
Explain methods that improve software reliability, focusing on techniques for finding and fixing errors in code.
Prompt B (generalized)
Write about practices that enhance code quality, particularly approaches to identifying problems and ensuring correctness.
Prompt C (very broad)
(start) Write about techniques for improving code quality and reliability. (end)

Example 10
Prompt A (highly abstracted)
Write about how different approaches to problem solving affect outcomes in technology and society.
Prompt B (highly abstracted)
Explain how methodologies shape the way humans build systems and solve challenges.
Prompt C (extremely broad)
(start) Write about how approaches influence outcomes. (end)

Example 11
Prompt A (highly abstracted)
Discuss the role of emerging technologies in reshaping human activity, including benefits and challenges.
Prompt B (highly abstracted)
Write about how innovations impact the way people live and work, considering risks and opportunities.
Prompt C (extremely broad)
(start) Write about how technology influences human activity. (end)

Example 12
Prompt A (highly abstracted)
Explain how humans improve the systems they create to make them more effective and reliable.
Prompt B (highly abstracted)
Write about methods for refining processes to achieve better performance and consistency.
Prompt C (extremely broad)
(start) Write about how humans refine systems to improve them. (end)

Example 13
Prompt A:
Write a poem that reflects on natural landscapes, using imagery and metaphor to highlight the relationship between humans and the environment.
Prompt B:
Compose a poem exploring the power of nature, focusing on themes of transformation, resilience, and human connection to the earth.
Prompt C:
(start)Poem on Nature(end)

Example 14
Prompt A:
Write an essay examining how revolutions throughout history have transformed societies, paying attention to causes, consequences, and cultural shifts.
Prompt B:
Compose a structured essay analyzing historical uprisings, their impact on leadership, governance, and social identity.
Prompt C:
(start)Essay on Revolutions(end)

Example 15
Prompt A:
Write a program that processes user input, applies logical conditions, and outputs meaningful results, ensuring clear structure and readability.
Prompt B:
Compose code that implements basic algorithms with input handling, decision-making, and structured output, emphasizing clarity and functionality.
Prompt C:
(start)Code on Input Processing(end)

Now carry out this task and output a prompt C with the following prompts A & B:
Prompt A: {story1}
Prompt B: {story2} 

\end{lstlisting}
\end{myprop}

%% file: appendix/experiment_details.tex
\section{Experiment Details}
\label{sec:exp_details}

We adopt the original implementation of EvalTree at \url{https://github.com/zhiyuan-zeng/evaltree} and use their default hyperparameters except for setting $k=2$. 
We report our LLM-as-a-judge prompt for faithfulness in  Template \ref{prompt:faith_prompt} and our prompt for informativeness in Template \ref{prompt:info_prompt}.

\begin{figure}
    \centering
    \includegraphics[width=0.5\linewidth]{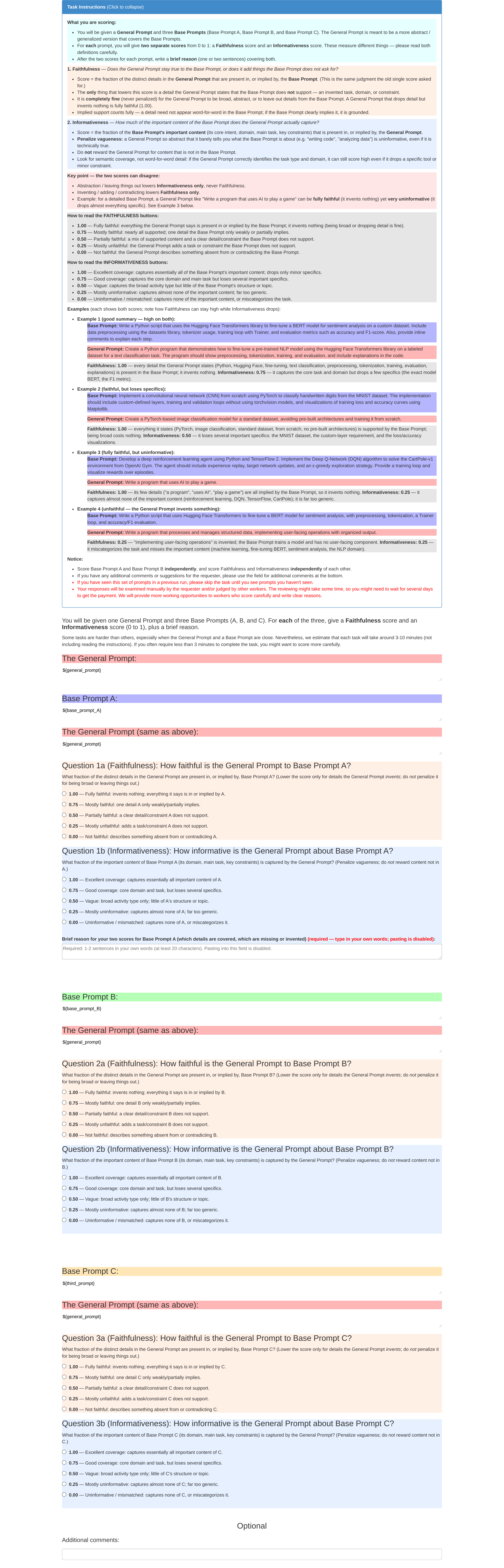}
    \caption{Our MTurk task template}
    \label{fig:MTurk template}
\end{figure}

\subsection{Human Experiments}

We sample $30$ common summaries from our entailment hierarchy. Our MTurk task instruction and questions can be seen in \Cref{fig:MTurk template}. The order of A, B, and C are randomized in the task. In the template, we ask the workers to justify their scores for the first prompt pair and reject the responses without a reason. We set the reward of each task to be \$1 - \$1.3 to make their hourly wage more than \$12. Among $111$ annotated tasks from $53$ workers, Claude filters $23$ tasks from $3$ workers and further suggests $29$ borderline annotations from $2$ workers.



\subsection{Agentic Evaluation}

Judging which hierarchy could help researchers to find weaknesses is challenging and subjective. Getting stable human evaluation for this application requires recruiting a sufficient number of experts, which is very expensive and whose results are hard to reproduce. So we use LLM agent as a human proxy to write the performance report. 

We choose to analyze \texttt{gemma-7b-it} and \texttt{Llama-2-7b-chat-hf} because they are famous, their average response scores are similar (5.5625 for Gemma and 5.5156 for Llama 2), and they have more obvious weaknesses than most of other models in WildBench.



\begin{myprop}\label{prompt:faith_prompt}\begin{lstlisting}
Faithfulness Evaluation Prompt (Posthoc)
System message:
You are a professional editor. Read the following prompt carefully and respond to the best of your ability.
User message:
You will be given a prompt A (child) and a general prompt B (parent) that is meant to be a generalization of prompt A. Your task is to determine how FAITHFUL prompt B is to prompt A.
Faithfulness measures whether everything prompt B asks for is actually grounded in prompt A. Calculate a score between 0 and 1 for what fraction of the distinct details in prompt B (the parent) are present in, or implied by, prompt A (the child).
Important:
- This is a check for the ABSENCE of hallucination. A faithful parent introduces no task, domain, constraint, or detail that prompt A does not support.
- Do NOT penalize abstraction or omission. It is expected and correct for B to drop specifics from A (e.g., specific libraries, datasets, or minor constraints) — even to keep only A's topic or domain and drop A's task or format. Leaving things out never lowers faithfulness — a parent that drops detail but invents nothing is fully faithful (1.0) — and "B omits A's requirements" is an informativeness concern, not grounds to lower this score. This protects only what B keeps: B is still penalized as usual for any detail it states that A does not support.
- ONLY penalize content in B that does not come from A — constraints B invents, or a task/domain B assigns that A is not actually about.
- Implicit support counts fully. A detail in B does NOT need to appear verbatim in A — if A clearly implies it, treat it as supported and do not deduct for it. Never lower the score just because A does not state B's wording explicitly. An abstract summary whose details are all implied (not stated) by A is well-grounded (0.6 and above), not a mismatch.
- Implication can be partial. If a detail in B is typically but not necessarily implied by A (it usually follows but is not guaranteed), give it partial credit rather than full credit, rather than treating it as fully grounded.
- A very broad or vague parent can still be faithful, as long as the little it does say is supported by the child.
Base your score on the ratio of B's distinct details that are supported by A, out of the total distinct details in B.
Scoring guidance:
- 0.85-1.0 (Fully Grounded): Every detail, task, and constraint in Prompt B is present in or directly implied by Prompt A. B may be far more abstract than A, but it adds nothing A does not support.
- 0.6-0.85 (Mostly Grounded): Nearly all of B is supported by A — explicitly or implicitly — with only a minor unsupported nuance, a slight shift in framing, or a detail that is typically but not necessarily implied by A. An abstract summary whose details are all implied by A (even if never stated verbatim) belongs here.
- 0.3-0.6 (Partially Grounded): B mixes supported content with a noticeable detail or constraint that A does not support even implicitly. (Reserve this for genuinely absent or invented content — not for details that A merely implies rather than states.)
- 0.0-0.3 (Hallucinated / Mismatched): B introduces a task, domain, or constraint that is absent from or contradicts A (e.g., A trains an ML model, but B describes user-facing CRUD operations).
## Examples
Prompt A (child): Implement a convolutional neural network (CNN) from scratch using PyTorch to classify handwritten digits from MNIST, with custom layers, training/validation loops without torchvision.models, and loss/accuracy curves in Matplotlib. Prompt B (parent): Write a Python program to train a deep learning model for image classification. Score: 1.0 Reason: Every claim in B — Python, training a deep learning model, image classification — is fully present in A. B abstracts away CNNs, MNIST, and Matplotlib, but omission is never penalized, and B invents nothing, so faithfulness is perfect.
Prompt A (child): Develop a deep reinforcement learning agent. Implement the Deep Q-Network (DQN) algorithm to solve the CartPole-v1 environment from OpenAI Gym. Prompt B (parent): Writing code. Score: 0.75 Reason: B's only claim is "writing code," which is usually — but not strictly — implied by developing a reinforcement-learning agent (one could develop an agent conceptually, or with no-code/configuration tools). Because the detail is probable rather than guaranteed, it is mostly grounded rather than fully grounded. (B is also extremely vague, but that is an informativeness issue, not a faithfulness one.)
Prompt A (child): Write a Python web scraper using requests and BeautifulSoup that collects book titles and prices from an online bookstore's catalog pages and writes them to a CSV file. Prompt B (parent): Write a Python program that gathers information from a website and saves it for later use. Score: 0.7 Reason: None of B's wording appears verbatim in A, but A clearly implies all of it: "gathers information from a website" is exactly what the scraper does, and "saves it" is implied by writing to a CSV — implicit support counts, so these are grounded, not absent. "For later use" is a probable but not guaranteed purpose, so it earns partial rather than full credit. B is an abstract but well-grounded generalization, so it is mostly grounded — not the 0.3 a strict "not stated explicitly" reading would give.
Prompt A (child): Write a 1,000-word op-ed arguing that city councils should expand protected bike lanes, citing reduced traffic fatalities and lower carbon emissions. Prompt B (parent): Write an opinion piece that argues for a local policy change and backs it up with concrete benefits. Score: 0.65 Reason: Every detail of B is implied by A rather than stated outright: "opinion piece" covers an op-ed, "argues for a local policy change" is implied by advocating that city councils expand bike lanes, and "concrete benefits" is implied by citing fewer fatalities and lower emissions. Implicit support counts, so none of this is invented. Only a small penalty applies — for the abstraction and for "concrete benefits" being a loose generalization of A's specific reasons — leaving it mostly grounded rather than a 0.3.
Prompt A (child): Write a Python script that uses Hugging Face Transformers to fine-tune a BERT model for sentiment analysis, including preprocessing, tokenization, a Trainer training loop, and accuracy/F1 evaluation. Prompt B (parent): Write a program that processes and manages structured data, implementing user-facing operations with organized output. Score: 0.30 Reason: "Processes structured data" is loosely supported, but "implementing user-facing operations" is a hallucinated constraint — A trains a model and has no user-facing/CRUD component. The low score is for this invented task, NOT for B dropping A's specifics (BERT, sentiment analysis, F1) — that omission alone would not lower faithfulness.
## Prompt A (child)
{prompt_a}
## Prompt B (parent)
{prompt_b}
Output your score in the following format:
Score: [SCORE]
\end{lstlisting}
\end{myprop}

\begin{myprop}\label{prompt:info_prompt}\begin{lstlisting}
Informativeness Evaluation Prompt (Posthoc)
System message:
You are a professional editor. Read the following prompt carefully and respond to the best of your ability.
User message:
You will be given a prompt A (child) and a general prompt B (parent) that is meant to be a generalization of prompt A. Your task is to determine how INFORMATIVE prompt B is about prompt A.
Informativeness measures whether prompt B retains the important content of prompt A. Calculate a score between 0 and 1 for what fraction of the important contents of prompt A (the child) are present in, or implied by, prompt B (the parent).
Important:
- First identify the IMPORTANT contents of A: its core intent, domain, primary task, and key constraints. Ignore trivial or incidental details.
- We are looking for SEMANTIC COVERAGE, not strict detail preservation. If B correctly identifies the type of task and domain of A (e.g., "write an analytical essay about climate policy" or "code a data-processing script"), it should score highly even if it abstracts away specific tools or minor constraints.
- Penalize VAGUENESS: a parent so abstract that it gives almost no structural or topical guidance about A (e.g., "writing code", "analyzing data") is uninformative, even if technically true.
- Do NOT reward B for content not in A; only the important content of A that B actually captures counts.
Base your score on the ratio of A's important contents that are captured by B, out of A's total important contents.
Scoring guidance:
- 0.85-1.0 (Excellent Coverage): B captures essentially all of A's important content — its domain, primary task, and key constraints — dropping only minor specifics (a particular library, dataset, or metric).
- 0.6-0.85 (Good Coverage): B captures A's core domain and primary task but loses several important specifics or constraints (e.g., keeps "train a deep learning model for image classification" but drops "from scratch", the dataset, and the visualization requirement).
- 0.3-0.6 (Vague / Gerund Phrase): B is extremely abstract (e.g., "Analyzing data", "Writing content"). It captures the broad activity type but almost none of A's structural or topical content.
- 0.0-0.3 (Uninformative / Mismatched): B captures essentially none of A's important content, because it is far too generic or it miscategorizes the task.
## Examples
Prompt A (child): Implement a convolutional neural network (CNN) from scratch using PyTorch to classify handwritten digits from MNIST, with custom layers, training/validation loops without torchvision.models, and loss/accuracy curves in Matplotlib. Prompt B (parent): Write a Python program to train a deep learning model for image classification. Score: 0.75 Reasoning (for your understanding only): B captures A's core content — Python, training a deep learning model, image classification — but loses several important specifics: the "from scratch" constraint, the MNIST dataset, and the visualization requirement. Good coverage of the core, not excellent.
Prompt A (child): Develop a deep reinforcement learning agent using Python and TensorFlow 2. Implement the Deep Q-Network (DQN) algorithm to solve the CartPole-v1 environment from OpenAI Gym. Prompt B (parent): Writing code. Score: 0.15 Reasoning (for your understanding only): B conveys none of A's important content — not the reinforcement-learning domain, the DQN algorithm, TensorFlow, or CartPole. Technically true, but uninformative.
Prompt A (child): Write a Python script that uses Hugging Face Transformers to fine-tune a BERT model for sentiment analysis, including preprocessing, tokenization, a Trainer training loop, and accuracy/F1 evaluation. Prompt B (parent): Write a program that processes and manages structured data, implementing user-facing operations with organized output. Score: 0.15 Reasoning (for your understanding only): B miscategorizes A. It captures none of A's important content — machine learning, fine-tuning BERT, sentiment analysis, the NLP domain — describing a different kind of program entirely.
## Prompt A (child)
{prompt_a}
## Prompt B (parent)
{prompt_b}
Output your score in the following format:
Score: [SCORE]
\end{lstlisting}
\end{myprop}

\section{AI Usage}
We use Claude Code to generate some analysis code and MTurk templates. We also use Claude and ChatGPT to help us search for some related work, or provide writing suggestions.

%% file: custom.bib
@article{ivison2024unpacking,
  title={Unpacking dpo and ppo: Disentangling best practices for learning from preference feedback},
  author={Ivison, Hamish and Wang, Yizhong and Liu, Jiacheng and Wu, Zeqiu and Pyatkin, Valentina and Lambert, Nathan and Smith, Noah A and Choi, Yejin and Hajishirzi, Hannaneh},
  journal={Advances in neural information processing systems},
  volume={37},
  pages={36602--36633},
  year={2024}
}

@inproceedings{ultrafeedback,
author = {Cui, Ganqu and Yuan, Lifan and Ding, Ning and Yao, Guanming and He, Bingxiang and Zhu, Wei and Ni, Yuan and Xie, Guotong and Xie, Ruobing and Lin, Yankai and Liu, Zhiyuan and Sun, Maosong},
title = {ULTRAFEEDBACK: boosting language models with scaled AI feedback},
year = {2024},
publisher = {JMLR.org},
booktitle = {Proceedings of the 41st International Conference on Machine Learning},
articleno = {384},
numpages = {23},
location = {Vienna, Austria},
series = {ICML'24}
}

@inproceedings{gabow1990data,
  title={Data structures for weighted matching and nearest common ancestors with linking},
  author={Gabow, Harold N},
  booktitle={Proceedings of the first annual ACM-SIAM symposium on Discrete algorithms},
  pages={434--443},
  year={1990}
}

@inproceedings{lin2025wildbench,
  title={Wildbench: Benchmarking llms with challenging tasks from real users in the wild},
  author={Lin, Bill Yuchen and Deng, Yuntian and Chandu, Khyathi and Ravichander, Abhilasha and Pyatkin, Valentina and Dziri, Nouha and Le Bras, Ronan and Choi, Yejin},
  booktitle={International Conference on Learning Representations},
  volume={2025},
  pages={47852--47870},
  year={2025}
}

@article{zhang2026skill,
  title={Skill-Aware Data Selection and Fine-Tuning for Data-Efficient Reasoning Distillation},
  author={Zhang, Lechen and Zhang, Yunxiang and Hu, Wei and Wang, Lu},
  journal={arXiv preprint arXiv:2601.10109},
  year={2026}
}

@article{okamoto2026trust,
  title={Trust by Design: Skill Profiles for Transparent, Cost-Aware LLM Routing},
  author={Okamoto, Mika and Erol, Ansel Kaplan and Matlin, Glenn},
  journal={arXiv preprint arXiv:2602.02386},
  year={2026}
}

@inproceedings{wang2026scan,
  title={SCAN: Structured Capability Assessment and Navigation for LLMs},
  author={Wang, Zongqi and Gu, Tianle and Gong, Chen and Tian, Xin and Bao, Siqi and Yang, Yujiu},
  booktitle={Proceedings of the 64th Annual Meeting of the Association for Computational Linguistics (Volume 1: Long Papers)},
  pages={2760--2799},
  year={2026}
}

@inproceedings{kim2026evalet,
  title={Evalet: Evaluating large language models through functional fragmentation},
  author={Kim, Tae Soo and Lee, Heechan and Lee, Yoonjoo and Seering, Joseph and Kim, Juho},
  booktitle={Proceedings of the 2026 CHI Conference on Human Factors in Computing Systems},
  pages={1--27},
  year={2026}
}

@article{bhuiya2026prompt2box,
  title={PROMPT2BOX: Uncovering Entailment Structure among LLM Prompts},
  author={Bhuiya, Neeladri and Dasgupta, Shib Sankar and McCallum, Andrew and Chang, Haw-Shiuan},
  journal={arXiv preprint arXiv:2603.21438},
  year={2026}
}

@inproceedings{reimers2019sentence,
  title={Sentence-bert: Sentence embeddings using siamese bert-networks},
  author={Reimers, Nils and Gurevych, Iryna},
  booktitle={Proceedings of the 2019 conference on empirical methods in natural language processing and the 9th international joint conference on natural language processing (EMNLP-IJCNLP)},
  pages={3982--3992},
  year={2019}
}

@article{frick2025prompt,
  title={Prompt-to-leaderboard},
  author={Frick, Evan and Chen, Connor and Tennyson, Joseph and Li, Tianle and Chiang, Wei-Lin and Angelopoulos, Anastasios N and Stoica, Ion},
  journal={arXiv preprint arXiv:2502.14855},
  year={2025}
}

@inproceedings{tjuatja2025behaviorbox,
  title={BehaviorBox: Automated discovery of fine-grained performance differences between language models},
  author={Tjuatja, Lindia and Neubig, Graham},
  booktitle={Proceedings of the 63rd Annual Meeting of the Association for Computational Linguistics (Volume 1: Long Papers)},
  pages={18851--18873},
  year={2025}
}

@inproceedings{huang2026probellm,
  title={ProbeLLM: Automating Principled Diagnosis of LLM Failures},
  author={Huang, Yue and Jiang, Zhengzhe and Ma, Yuchen and Jiang, Yu and Wang, Xiangqi and Zhou, Yujun and Hao, Yuexing and Guo, Kehan and Chen, Pin-Yu and Ghassemi, Marzyeh and others},
  booktitle={International Conference on Machine Learning},
  year={2026}
}

@article{ashury2026errormap,
  title={ErrorMap and ErrorAtlas: Charting the Failure Landscape of Large Language Models},
  author={Ashury-Tahan, Shir and Mai, Yifan and Bandel, Elron and Shmueli-Scheuer, Michal and Choshen, Leshem},
  journal={arXiv preprint arXiv:2601.15812},
  year={2026}
}

@inproceedings{cheng2024autodetect,
  title={Autodetect: Towards a unified framework for automated weakness detection in large language models},
  author={Cheng, Jiale and Lu, Yida and Gu, Xiaotao and Ke, Pei and Liu, Xiao and Dong, Yuxiao and Wang, Hongning and Tang, Jie and Huang, Minlie},
  booktitle={Findings of the Association for Computational Linguistics: EMNLP 2024},
  pages={6786--6803},
  year={2024}
}

@inproceedings{sanders2024tv,
  title={Tv-trees: Multimodal entailment trees for neuro-symbolic video reasoning},
  author={Sanders, Kate and Weir, Nathaniel and Van Durme, Benjamin},
  booktitle={Proceedings of the 2024 Conference on Empirical Methods in Natural Language Processing},
  pages={19009--19028},
  year={2024}
}

@inproceedings{weir2024nellie,
  title={NELLIE: a neuro-symbolic inference engine for grounded, compositional, and explainable reasoning},
  author={Weir, Nathaniel and Clark, Peter and Van Durme, Benjamin},
  booktitle={Proceedings of the Thirty-Third International Joint Conference on Artificial Intelligence},
  pages={3602--3612},
  year={2024}
}

@article{huang2025retrieval,
  title={Retrieval-Augmented Generation with Hierarchical Knowledge},
  author={Huang, Haoyu and Huang, Yongfeng and Yang, Junjie and Pan, Zhenyu and Chen, Yongqiang and Ma, Kaili and Chen, Hongzhi and Cheng, James},
  journal={arXiv preprint arXiv:2503.10150},
  year={2025}
}

@article{gao2025science,
  title={Science hierarchography: Hierarchical organization of science literature},
  author={Gao, Muhan and Shah, Jash and Wang, Weiqi and Huang, Kuan-Hao and Khashabi, Daniel},
  journal={arXiv preprint arXiv:2504.13834},
  year={2025}
}

@inproceedings{zhong2025hicode,
  title={HICode: Hierarchical inductive coding with LLMs},
  author={Zhong, Mian and Wang, Pristina and Field, Anjalie},
  booktitle={Proceedings of the 2025 Conference on Empirical Methods in Natural Language Processing},
  pages={31048--31066},
  year={2025}
}

@article{hsu2024chime,
  title={Chime: Llm-assisted hierarchical organization of scientific studies for literature review support},
  author={Hsu, Chao-Chun and Bransom, Erin and Sparks, Jenna and Kuehl, Bailey and Tan, Chenhao and Wadden, David and Wang, Lucy Lu and Naik, Aakanksha},
  journal={Findings of the Association for Computational Linguistics: ACL 2024},
  pages={118--132},
  year={2024}
}

@article{gupta2025llm,
  title={LLM-guided Hierarchical Retrieval},
  author={Gupta, Nilesh and Chang, Wei-Cheng and Bui, Ngot and Hsieh, Cho-Jui and Dhillon, Inderjit S},
  journal={arXiv preprint arXiv:2510.13217},
  year={2025}
}

@inproceedings{kargupta2025taxoadapt,
  title={Taxoadapt: Aligning llm-based multidimensional taxonomy construction to evolving research corpora},
  author={Kargupta, Priyanka and Zhang, Nan and Zhang, Yunyi and Zhang, Rui and Mitra, Prasenjit and Han, Jiawei},
  booktitle={Proceedings of the 63rd Annual Meeting of the Association for Computational Linguistics (Volume 1: Long Papers)},
  pages={29834--29850},
  year={2025}
}

@inproceedings{zhang2025sirerag,
  title={Sirerag: Indexing similar and related information for multihop reasoning},
  author={Zhang, Nan and Choubey, Prafulla Kumar and Fabbri, Alexander and Bernadett-Shapiro, Gabriel and Zhang, Rui and Mitra, Prasenjit and Xiong, Caiming and Wu, Chien-Sheng},
  booktitle={International Conference on Learning Representations},
  volume={2025},
  pages={100413--100432},
  year={2025}
}

@inproceedings{sarthi2024raptor,
  title={Raptor: Recursive abstractive processing for tree-organized retrieval},
  author={Sarthi, Parth and Abdullah, Salman and Tuli, Aditi and Khanna, Shubh and Goldie, Anna and Manning, Christopher},
  booktitle={International Conference on Learning Representations},
  volume={2024},
  pages={32628--32649},
  year={2024}
}

@inproceedings{tian2024generic,
  title={A generic method for fine-grained category discovery in natural language texts},
  author={Tian, Chang and Blaschko, Matthew B and Yin, Wenpeng and Xing, Mingzhe and Yue, Yinliang and Moens, Marie Francine},
  booktitle={Proceedings of the 2024 Conference on Empirical Methods in Natural Language Processing},
  pages={3548--3566},
  year={2024}
}

@inproceedings{zhang2025llmtaxo,
  title={LLMTaxo: Leveraging Large Language Models for Constructing Taxonomy of Factual Claims from Social Media},
  author={Zhang, Haiqi and Zhu, Zhengyuan and Zhang, Zeyu and Li, Chengkai},
  booktitle={Findings of the Association for Computational Linguistics: ACL 2025},
  pages={19627--19641},
  year={2025}
}

@inproceedings{cattan2023champ,
  title={CHAMP: Efficient Annotation and Consolidation of Cluster Hierarchies},
  author={Cattan, Arie and Hope, Tom and Downey, Doug and Bar-Haim, Roy and Eden, Lilach and Kantor, Yoav and Dagan, Ido},
  booktitle={Proceedings of the 2023 Conference on Empirical Methods in Natural Language Processing: System Demonstrations},
  pages={403--412},
  year={2023}
}

@article{jiang2026artificial,
  title={Artificial hivemind: The open-ended homogeneity of language models (and beyond)},
  author={Jiang, Liwei and Chai, Yuanjun and Li, Margaret and Liu, Mickel and Fok, Raymond and Dziri, Nouha and Tsvetkov, Yulia and Sap, Maarten and Choi, Yejin},
  journal={Advances in Neural Information Processing Systems},
  volume={38},
  year={2026}
}

@inproceedings{budagam2025hierarchical,
  title={Hierarchical Prompting Taxonomy: A Universal Evaluation Framework for Large Language Models Aligned with Human Cognitive Principles},
  author={Budagam, Devichand and Kumar, Ashutosh and Khoshnoodi, Mahsa and KJ, Sankalp and Jain, Vinija and Chadha, Aman},
  booktitle={First International KDD Workshop on Prompt Optimization},
  year={2025}
}

@article{zhao2026hierarchical,
  title={Hierarchical Abstract Tree for Cross-Document Retrieval-Augmented Generation},
  author={Zhao, Ziwen and Yang, Menglin},
  journal={arXiv preprint arXiv:2605.00529},
  year={2026}
}

@inproceedings{dalvi2021explaining,
  title={Explaining Answers with Entailment Trees},
  author={Dalvi, Bhavana and Jansen, Peter and Tafjord, Oyvind and Xie, Zhengnan and Smith, Hannah and Pipatanangkura, Leighanna and Clark, Peter},
  booktitle={Proceedings of the 2021 Conference on Empirical Methods in Natural Language Processing},
  pages={7358--7370},
  year={2021}
}

@article{chen2023walking,
  title={Walking down the memory maze: Beyond context limit through interactive reading},
  author={Chen, Howard and Pasunuru, Ramakanth and Weston, Jason and Celikyilmaz, Asli},
  journal={arXiv preprint arXiv:2310.05029},
  year={2023}
}

@article{tamkin2024clio,
  title={Clio: Privacy-Preserving Insights into Real-World AI Use},
  author={Tamkin, Alex and McCain, Miles and Handa, Kunal and Durmus, Esin and Lovitt, Liane and Rathi, Ankur and Huang, Saffron and Mountfield, Alfred and Hong, Jerry and Ritchie, Stuart and others},
  journal={arXiv preprint arXiv:2412.13678},
  year={2024}
}

@article{tian2025skillverse,
  title={SkillVerse: Assessing and Enhancing LLMs with Tree Evaluation},
  author={Tian, Yufei and Sun, Jiao and Peng, Nanyun and Zhang, Zizhao},
  journal={arXiv preprint arXiv:2506.00319},
  year={2025}
}

@article{zeng2025evaltree,
  title={Evaltree: Profiling language model weaknesses via hierarchical capability trees},
  author={Zeng, Zhiyuan and Wang, Yizhong and Hajishirzi, Hannaneh and Koh, Pang Wei},
  journal={arXiv preprint arXiv:2503.08893},
  year={2025}
}

@inproceedings{chang2024fine,
  title={Fine-to-coarse entailment hierarchy construction for coarse-to-fine story generation},
  author={Chang, Haw-Shiuan and Peng, Nanyun and Bansal, Mohit and Chung, Tagyoung},
  booktitle = {Bridging Human-Computer Interaction and Natural Language Processing (HCI+NLP)},
  year={2024}
}
